\documentclass[sigconf,10pt]{acmart}
\renewcommand\footnotetextcopyrightpermission[1]{}

\usepackage{tikz}
\usepackage{amsmath}

\usepackage{filecontents}

\usepackage{cancel}
\usepackage{outlines}
\usepackage{enumitem}
\usepackage{listings}
\setlist[enumerate,2]{label=\roman*)}
\setlist[enumerate,3]{label=\alph*)}
\newcommand{\highlight}[2]{\colorbox{#1!17}{#2}}

\usepackage{tikz}
\usepackage{amsmath}
\usepackage{pifont}
\usepackage[english]{babel}
\usepackage{blindtext}
\usepackage{lipsum}

\usepackage{algorithm}
\usepackage{algpseudocode}

\algnewcommand\Var[1]{\ensuremath{\mathit{#1}}}

\usepackage{filecontents}
\usepackage{xspace}
\usepackage[justification=centering]{subfig}
\usepackage{xcolor}
\usepackage{hhline}
\usepackage{multirow}
\usepackage{soul}

\usepackage{array}
\usepackage{comment}

\copyrightyear{2023} 
\acmYear{2023} 
\setcopyright{acmlicensed}
\acmConference[SoCC '23]{ACM Symposium on Cloud Computing}{October 
30--November 1, 2023}{Santa Cruz, CA, USA}
\acmBooktitle{ACM Symposium on Cloud Computing (SoCC '23), October 
30--November 1, 2023, Santa Cruz, CA, USA}
\acmPrice{15.00}
\acmDOI{10.1145/3620678.3624653}
\acmISBN{979-8-4007-0387-4/23/10}

\begin{document}

\newcommand*{\affaddr}[1]{#1}
\newcommand*{\affmark}[1][*]{\textsuperscript{#1}}

\author{
{\large Kuntai Du, Yuhan Liu, Yitian Hao, Qizheng Zhang\affmark[$\ddag$],\\Haodong Wang, Yuyang Huang, Ganesh Ananthanarayanan\affmark[$\dag$], Junchen Jiang}\\
\affaddr{\textit{\large{University of Chicago}}}~~~~~\affaddr{\textit{\large \affmark[$\ddag$]Stanford University}}~~~~~\affaddr{\textit{\large \affmark[$\dag$]Microsoft Research}}\\
}

\renewcommand{\shortauthors}{K. Du, Y. Liu, Y. Hao, Q. Zhang, H. Wang, Y. Huang, G. Ananthanarayanan, J. Jiang}

\setcopyright{none}
\pagestyle{plain} 

\soulregister\sf7




\newcommand{\name}{\textsf{OneAdapt}\xspace}
\newcommand{\accchange}{Accuracy-knob Change\xspace}
\newcommand{\delaygrad}{Delay-knob Gradient\xspace}
\newcommand{\resgrad}{Resource-knob Gradient\xspace}
\newcommand{\outputgrad}{OutputGrad\xspace}
\newcommand{\accgrad}{{AccGrad}\xspace}
\newcommand{\accres}{Accuracy-resource Property\xspace}
\newcommand{\blackbox}{Black-box Property \xspace}
\newcommand{\accproxy}{Accuracy Change Proxy}
\newcommand{\Output}{\ensuremath{\textbf{z}}\xspace}
\newcommand{\Config}{\ensuremath{\textbf{k}}\xspace}
\newcommand{\Data}{\ensuremath{\textbf{x}}\xspace}
\newcommand{\Knob}{\ensuremath{k}\xspace}
\newcommand{\Input}{\ensuremath{\textbf{y}}\xspace}
\newcommand{\inputgrad}{InputGrad\xspace}
\newcommand{\dnngrad}{DNNGrad\xspace}
\newcommand{\summary}{output utility\xspace}
\newcommand{\Summary}{Output utility\xspace}
\newcommand{\spatial}{spatial\xspace}
\newcommand{\Spatial}{Spatial\xspace}
\newcommand{\Temporal}{Temporal\xspace}
\newcommand{\temporal}{temporal\xspace}
\newcommand{\Uniform}{Coarse-grained\xspace}
\newcommand{\uniform}{coarse-grained\xspace}
\newcommand{\nonuniform}{fine-grained\xspace}
\newcommand{\Nonuniform}{Fine-grained\xspace}
\newcommand{\Resource}{\ensuremath{\textbf{r}}\xspace}
\newcommand{\Acc}{\ensuremath{Acc}\xspace}
\newcommand{\Grad}{\ensuremath{\textbf{g}}\xspace}

\newcommand{\cameranew}[1]{{{#1}\xspace}}
\newcommand{\cameraold}[1]{{\color{red}{}}}
\definecolor{kellygreen}{rgb}{0.3, 0.73, 0.09}
\definecolor{olivedrab}{rgb}{0.42,0.56,0.13}
\newcommand{\yt}[1]{{\color{olivedrab}{\footnotesize [YT: #1]\xspace}}}
\newcommand{\hd}[1]{{\color{kellygreen}{\footnotesize [HD: #1]\xspace}}}
\newcommand{\yh}[1]{{\color{violet}{\footnotesize [YH: #1]\xspace}}}
\newcommand{\qz}[1]{{\color{orange}{\footnotesize [QZ: #1]\xspace}}}
\newcommand{\jc}[1]{{\color{red}{\footnotesize [JC: #1]\xspace}}}
\newcommand{\kt}[1]{{\color{brown}{#1}}}
\newcommand{\roy}[1]{{\color{brown}{\footnotesize [ROY: #1]\xspace}}}
\newcommand{\ga}[1]{{\color{magenta}{\footnotesize [GA: #1]\xspace}}}
\newcommand{\edit}[1]{{\color{blue}{#1}}}
\newcommand{\hh}[1]{{\color{purple}{\footnotesize [HH: #1]}}}
\newcommand{\ap}[1]{{\color{purple}{\footnotesize [AP: #1]}}}
\newcommand{\ignore}[1]{{\xspace}}
\newcommand{\REMOVE}[1]{}

\newcommand{\review}[2]{{#1}}
\newcommand{\reviewreview}[2]{{#1}}

\newcommand{\fillme}{{\bf XXX}\xspace}

\newcommand*\circled[1]{\raisebox{.5pt}{\textcircled{\raisebox{-.9pt} {#1}}} }

\newcounter{packednmbr}
\newenvironment{packedenumerate}{\begin{list}{\thepackednmbr.}{\usecounter{packednmbr}\setlength{\itemsep}{0.5pt}\addtolength{\labelwidth}{-4pt}\setlength{\leftmargin}{2ex}\setlength{\listparindent}{\parindent}\setlength{\parsep}{1pt}\setlength{\topsep}{0pt}}}{\end{list}}
\newenvironment{packeditemize}{\begin{list}{$\bullet$}{\setlength{\itemsep}{0.5pt}\addtolength{\labelwidth}{-4pt}\setlength{\leftmargin}{2ex}\setlength{\listparindent}{\parindent}\setlength{\parsep}{1pt}\setlength{\topsep}{2pt}}}{\end{list}}
\newenvironment{packedpackeditemize}{\begin{list}{$\bullet$}{\setlength{\itemsep}{0.5pt}\addtolength{\labelwidth}{-4pt}\setlength{\leftmargin}{\labelwidth}\setlength{\listparindent}{\parindent}\setlength{\parsep}{1pt}\setlength{\topsep}{0pt}}}{\end{list}}
\newenvironment{packedtrivlist}{\begin{list}{\setlength{\itemsep}{0.2pt}\addtolength{\labelwidth}{-4pt}\setlength{\leftmargin}{\labelwidth}\setlength{\listparindent}{\parindent}\setlength{\parsep}{1pt}\setlength{\topsep}{0pt}}}{\end{list}}
\let\enumerate\packedenumerate
\let\endenumerate\endpackedenumerate
\let\itemize\packeditemize
\let\enditemize\endpackeditemize

\newcommand{\tightcaption}[1]{
\vspace{-0.1cm}
\caption{{\normalfont{\textit{{#1}}}}}
\vspace{-0.3cm}
}
\newcommand{\tightsection}[1]{\vspace{-0.1cm}\section{#1}\vspace{-0.05cm}}
\newcommand{\tightsectionstar}[1]{\vspace{-0.17cm}\section*{#1}\vspace{-0.08cm}}
\newcommand{\tightsubsection}[1]{\vspace{-0.3cm}\subsection{#1}\vspace{-0.1cm}}
\newcommand{\tightsubsubsection}[1]{\vspace{-0.01in}\subsubsection{#1}\vspace{-0.01cm}}

\newcommand{\eg}{{\it e.g.,}\xspace}
\newcommand{\ie}{{\it i.e.,}\xspace}
\newcommand{\etal}{{\it et.~al}\xspace}
\newcommand{\bigO}{\mathrm{O}}
\newcommand{\twlog}{w.l.o.g.\xspace}

\newcommand{\myparashort}[1]{\vspace{0.05cm}\noindent{\bf {#1}}~}
\newcommand{\mypara}[1]{\vspace{0.05cm}\noindent{\bf {#1}:}~}
\newcommand{\myparatight}[1]{\vspace{0.02cm}\noindent{\bf {#1}:}~}
\newcommand{\myparaq}[1]{\smallskip\noindent{\bf {#1}?}~}
\newcommand{\myparaittight}[1]{\smallskip\noindent{\emph {#1}:}~}
\newcommand{\question}[1]{\smallskip\noindent{\emph{Q:~#1}}\smallskip}
\newcommand{\myparaqtight}[1]{\smallskip\noindent{\bf {#1}}~}

\newcommand{\cmark}{\ding{51}}%
\newcommand{\xmark}{\ding{55}}%

\title{\name: Fast Configuration Adaptation for Video Analytics Applications via Backpropagation}

\begin{abstract}

Deep learning inference on streaming media data, such as object detection in video or LiDAR feeds and text extraction from audio waves, is now ubiquitous.
To achieve high inference accuracy, these applications typically require significant network bandwidth to gather high-fidelity data and extensive GPU resources to run deep neural networks (DNNs).
While the high demand for network bandwidth and GPU resources {\em could} be substantially reduced by {\em optimally adapting} the configuration knobs, such as video resolution and frame rate, current adaptation techniques fail to meet three requirements simultaneously:
adapt configurations {\em (i)} with minimum extra GPU or bandwidth overhead
{\em (ii)} to reach near-optimal decisions based on how the data affects the final DNN's accuracy, 
and {\em (iii)} do so for a range of configuration knobs.
This paper presents \name, which meets these requirements by leveraging a gradient-ascent strategy to adapt configuration knobs.
The key idea is to embrace DNNs' {\em differentiability} to quickly estimate the accuracy's gradient to each configuration knob, called \accgrad.
Specifically, \name estimates \accgrad by multiplying two gradients: 
\inputgrad (\ie how each configuration knob affects the input to the DNN) and \dnngrad (\ie how the DNN input affects the DNN inference output). 
We evaluate \name across five types of configurations, four analytic tasks, and five types of input data.
Compared to state-of-the-art adaptation schemes, \name cuts bandwidth usage and GPU usage by 15-59\% while maintaining comparable accuracy or improves accuracy by 1-5\%  while using equal or fewer resources.

\end{abstract}
\maketitle




\tightsection{Introduction}
\label{sec:intro}

\review{
Many real-world applications run deep neural networks (DNNs) to perform analytics on streaming media data, such as RGB videos, LiDAR point clouds, depth videos, and audio waves.
For example, autonomous-driving applications rely on DNNs to detect vehicles in individual video RGB frames and LiDAR point clouds~\cite{tusimple,waymo,trafficvision, traffictechnologytoday,goodvision,intuvisiontech, vision-zero, msr,servai-1,servai-2,servai-3}.
Similarly, smart-home applications use DNNs to extract text from audio segments~\cite{smarthome-1,smarthome-2,smarthome-3}.
We focus on these applications, which we refer to as \textit{\textbf{streaming media analytics}} (\S\ref{subsec:streaming-analytics}).
Notably, streaming media analytics encompass the popular video analytics applications, but not all DNN-based tasks (\eg generative tasks).
}{This addresses Reviewer C's comments on the scope of \name}



Many streaming media analytics systems can be resource-intensive, in \textbf{\textit{network bandwidth}} or \textit{\textbf{GPU cycles}} or both.
To achieve high accuracy, they run complex DNN inference on each frame (or segment) and require data at high fidelity, which potentially requires high GPU usage and network bandwidth (if a remote sensor captures the data). 


To reduce resource usage, many prior solutions (\eg~\cite{dds,eaar,elf,vigil,glimpse,accmpeg,reducto,casva}) apply 
\textit{input filtering} to downsample or drop redundant data and then run
\textit{DNN inference} to analyze the filtered input. 
Ideally, if the input filtering is configured with an optimal setting for its {\em knobs} (\eg video frame rate and resolution), the bandwidth usage and GPU usage can be drastically reduced without affecting inference accuracy.
(Table~\ref{tab:knob-categorization} summarizes some popular filtering knobs.)

The challenge of optimally adapting the filtering knobs is that the optimal setting of these knobs \textit{\textbf{varies}} over time as input data evolves~\cite{chameleon,awstream,vstore,casva,videostorm,reducto}.
We call the optimal setting of these knobs an {\em optimal configuration}.
For example, when vehicles stop at the red light, feeding the traffic video at a low frame rate to a DNN can still accurately detect vehicles,
but when vehicles start moving fast, the DNN must run more frequently (\eg 30fps) to detect the vehicles.
In this case, the frame rate
must be adapted over time.

To handle this challenge, an ideal adaptation logic should meet three requirements:
\begin{packeditemize}
\item {\em (i)} {\em Frequent:} 
The adaptation logic can run frequently with minimum GPU computation and bandwidth overhead.
\item {\em (ii)} {\em Near-optimal:} 
The adaptation logic can pick a near-optimal configuration based on how various filtering on the data will affect the output of a particular final DNN.
\item {\em (iii)} {\em Generic:} 
The same logic can be applied to a wide range of analytic tasks, streaming media, and filtering knobs.
\end{packeditemize}





\begin{figure}
    \centering
    \includegraphics[width=0.8\columnwidth]{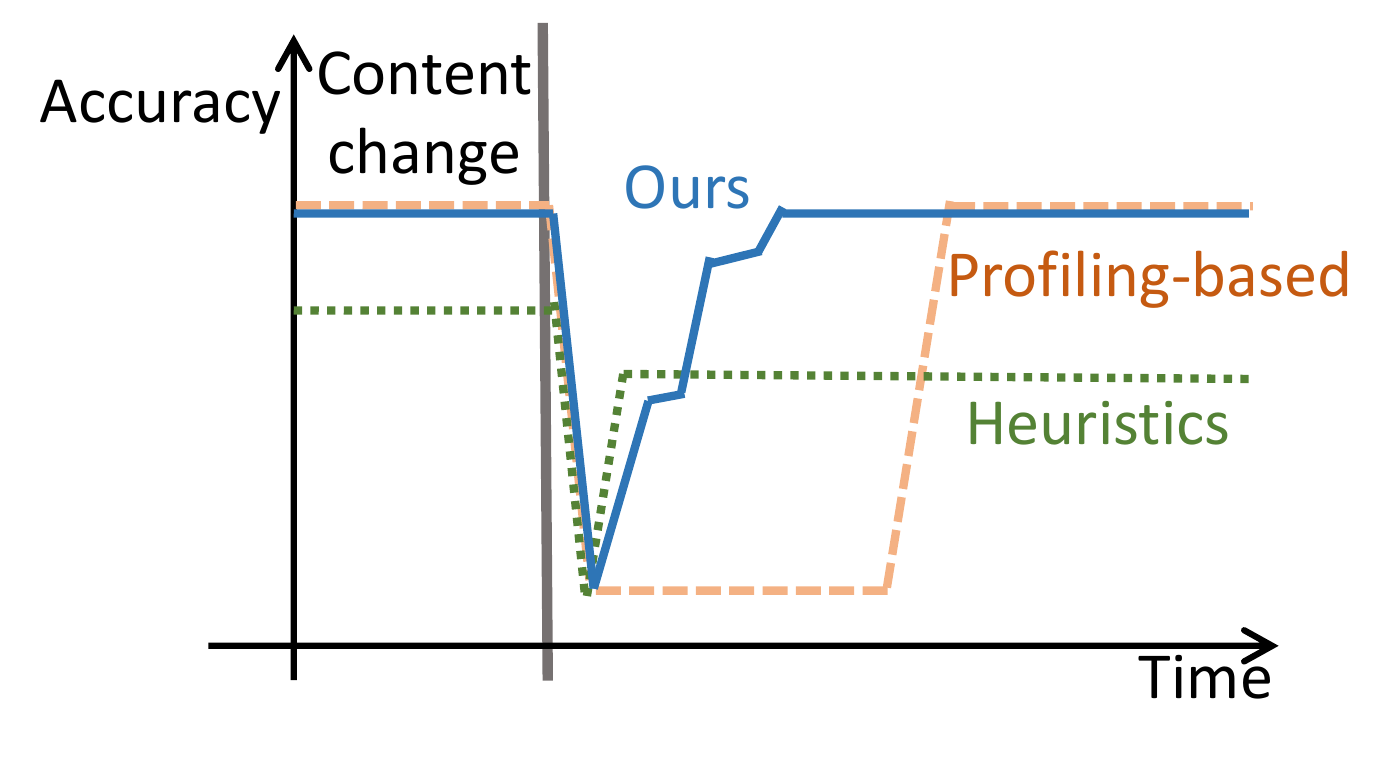}
    \vspace{-0.1cm}
    \tightcaption{
    Demonstrating the adaptive behavior of \name compared to alternatives: \name quickly adapts to a near-optimal configuration after the change of input content, whereas the profiling-based method adapts slowly and the heuristic-based method adapts quickly but suboptimally.
    }
    \label{fig:adaptation-behavior}
    \vspace{-0.1cm}
\end{figure}

\mypara{Prior approaches}
Prior efforts fall short in at least one of these requirements (detailed discussion in \S\ref{subsec:configuration-adaptation}).
\begin{packeditemize}
\item {\em Profiling-based} methods (\eg~\cite{awstream,chameleon,videostorm,vstore}) periodically run {\em extra} DNN inferences to profile the accuracy of alternative configurations, which incur significant GPU overhead.
For example, AWStream~\cite{awstream} profiles different combinations of resolutions, frame rates, and quantization parameters. 
Even after \textit{downsampling} the combinations, such profiling still has up to 17$\times$ more GPU computation than regular DNN inference.
With limited GPU resources, this profiling overhead decelerates the adaptation of configuration, leading to outdated configuration when the content of input data varies over time (as shown in Figure~\ref{fig:adaptation-behavior}). 

\item Alternatively, many {\em heuristic-based} methods (\eg~\cite{dds,reducto,vigil,glimpse,eaar,accmpeg,casva,elf,wang2022minimizing}) avoid extra inference and instead adapt configurations more frequently using cheap heuristics.
These heuristics make simplified assumptions about which parts of the input are unimportant to the final DNN, resulting in low accuracy (as shown in Figure\ref{fig:adaptation-behavior}). 
For instance, some heuristics filter a new frame for DNN inference when its pixels differ from the last analyzed frame significantly (\eg~\cite{reducto,glimpse,rocket}), but the pixel differences can be on the background rather than any object of interest.

\item Moreover, many existing techniques are designed for {\em specific} analytic tasks, streaming media and/or filtering knobs.
For instance, DDS~\cite{dds} and STAC~\cite{stac} select which regions in a frame should be in high quality, but its logic cannot be directly extended to adapt non-video input data, such as LiDAR point cloud, depth map, or audio waves.

\end{packeditemize}

\myparashort{Our approach: Fast gradient-based adaptation.}
We present \name, a configuration adaptation system that improves on all three fronts.
\name uses gradient ascent to continuously tune the filtering knobs to improve the tradeoff between accuracy and resource usage. 
It uses a new technique to \textit{cheaply} approximate the change of DNN inference accuracy in response to a small change on each filtering knob, which we refer to as \textit{\textbf{\accgrad}}.\footnote{\review{Note that we define \accgrad as \textit{numerical} gradient rather than analytical gradient, which is used~\cite{pcc-vivace,ramazanov2011stability,chistyakov2015stability} when optimizing discrete system knobs.}{This addresses Reviewer A's comments on whether \accgrad is numerical gradient or analytical gradient}}


The fast approximation of \accgrad is based on the observation that filtering knobs influence the inference accuracy through their changes on the DNN's input. 
Thus, the \accgrad of a filtering knob 
can be decoupled into two parts, both of which can be computed with low overheads.
\begin{packedenumerate}
\item {\em \inputgrad:} How the DNN's input changes with each knob, which can be done on CPU without GPU compute.
\item {\em \dnngrad:} How the DNN's accuracy changes with its input, which can be obtained by a single DNN backpropagation.
\end{packedenumerate}
We further speed up the existing backpropagation operator by not computing the gradients on DNN weights (\S\ref{sec:cost} for more cost reduction techniques).
Moreover, because \dnngrad describes DNN's sensitivity to its input change, regardless of which knob causes the input change, \dnngrad can be computed \textit{\textbf{once}} and then be multiplied with the \inputgrad of different knobs to get their \accgrad.

Since \accgrad can be approximated with minimum extra GPU compute, \name updates the estimate of \accgrad frequently (\eg every second) and runs a \textit{\textbf{gradient-ascent}} strategy to update the configuration frequently in a way that maximally increases accuracy or reduces resource usage without hurting accuracy.
Like in other similar settings~\cite{pcc-vivace}, this \textit{\textbf{gradient-ascent}} strategy is likely to converge to a near-optimal configuration, because the configuration-accuracy relationships are mostly {\em concave}.\footnote{The concavity can be intuitively explained as:
if we increase the knob by a small amount (\eg increase frame rate), the gain in accuracy is more significant when the system is of low accuracy, but the gain diminishes when the system is already generating accurate inference results.
}
\S\ref{sec:design} offers formal and intuitive explanation why \accgrad can be decoupled (\S\ref{subsec:outputgrad}) and why \name converges (\S\ref{subsec:limitation}).

To put \name's contribution into perspective, using DNN gradient is not new in video analytics systems~\cite{stac,saliency_hotmobile}, but to the best of our knowledge, \name is the first to enable fast approximation of accuracy's gradient with respect to a range of popular filtering knobs.
As illustrated in Figure~\ref{fig:adaptation-behavior}, \name outperforms both profiling-based and heuristic-based approaches.
Unlike profiling-based methods, we estimate \accgrad and adapt more frequently (by default, every second), while profiling-based methods adapt configuration after the slow profiling finishes (\eg every minute~\cite{awstream}). 
Unlike heuristics-based methods which either analyze the input data or the outputs of the DNN (including intermediate outputs~\cite{dds,eaar,elf}), \name can converge to a {\em closer-to-optimal} configuration as \accgrad directly indicates how DNN accuracy varies with a change in the configuration. 


We evaluate \name on nine streaming-media analytics pipelines (Table~\ref{tab:pipeline}) covering four analytic tasks, five types of input data 
and five filtering knobs.
We compare \name with the latest profiling-based or heuristic-based schemes designed for individual knob types.
Our key results are: 
\begin{packedenumerate}
\item \name reaches similar accuracy while reducing bandwidth usage or GPU compute by 15-59\%. 
Alternatively, \name improves accuracy by 1-5\% without using more bandwidth usage or GPU usage compared to the baselines.
\item 
\review{Compared to a straightforward implementation, \name reduces the GPU computation overhead and GPU memory overhead of \accgrad estimation by 87\% and 12\%, keeping \name's extra computation overhead below 20\% of DNN inference.
}{This addresses reviewer C's concern about the speed of backpropagation.}
\item Unlike solutions that are designed for specific filtering knobs, \name achieves its improvement across all filtering knobs using the same adaptation logic. 
\end{packedenumerate}
That said, \name still has its limitations (\S\ref{sec:limitation_section}), such as not handling non-filtering knobs (\eg DNN selection) and tasks outside streaming media analytics (\eg generative tasks). 

\textbf{This work does not raise any ethical issues.}


\section{Background}

\vspace{0.2cm}

\tightsubsection{DNN-based streaming media analytics}
\label{subsec:streaming-analytics}

\review{
DNN-based analytics applications are ubiquitous~\cite{tusimple,waymo,trafficvision, traffictechnologytoday,goodvision,intuvisiontech, vision-zero, msr,smarthome-1,smarthome-2,smarthome-3}.
In this paper, we focus on streaming media analytics, where the analytic tasks (\eg detection and segmentation) are performed on streaming media (\eg RGB video, LiDAR video, audio, etc).
Streaming media analytics are widely used in applications like autonomous driving (which requires object detection on RGB and LiDAR video), human detection (which requires human detection on InfraRed video at night), and smart home (which requires word detection on audio data).
Note that streaming media analytics include popular video analytics applications though not all DNN-based applications (\eg generative AI).
}{This addresses Reviewer C's comments on the scope of \name}

\begin{figure}
    \centering
    \includegraphics[width=.99\columnwidth]{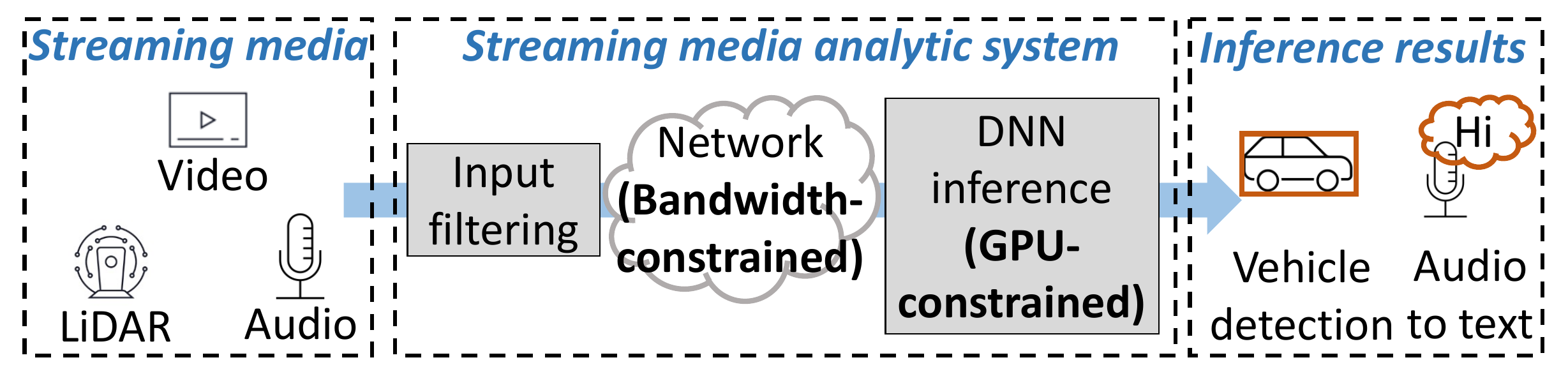}
    \vspace{-0.2cm}
    \tightcaption{
    In streaming media analytics, data from sensors is processed by the \textit{\textbf{input filtering}} module before being sent to the \textbf{\textit{DNN}} on constrained GPU resources. Given the DNN module might be on a remote device or cloud, data transmission can occur over a bandwidth-limited network.
    }
    \label{fig:dnn-analytics-application-flow}
\end{figure}

Typically, a streaming media analytic system, depicted in Figure~\ref{fig:dnn-analytics-application-flow}, consists of {\em input filtering} (typically at the sensor side) and {\em DNN inference} (typically in edge or cloud).
First, the system collects the raw data, referred to as the {\em input data}.
This data is then filtered by the input filtering process.
When the data needs to be sent to a separate analytics server (in edge or cloud), it will be streamed through a {\em bandwidth-constraint} link.
The server then runs DNN inference to get
inference results (\eg vehicle bounding boxes for autonomous driving) using \textit{limited} GPU resources.
Due to limited GPU resources and network bandwidth, not all data in their highest fidelity will be analyzed by the DNN.

\mypara{Objective}
The objective of these systems is to reduce resource usage (in GPU and bandwidth) without hurting the inference accuracy, or improve inference accuracy using limited resources.
Following prior work~\cite{awstream,chameleon,dds,accmpeg,ekya,yoda}, we define {\em inference accuracy} of an inference result as its similarity with the inference result obtained under unlimited bandwidth and/or GPU computation budget (see \S\ref{subsec:eval-setup} for detailed definition).
This definition of accuracy highlights the impact of saving resources on inference accuracy. 

\tightsubsection{Configuration adaptation}
\label{subsec:numer_of_knobs}

Prior work has shown that high accuracy {\em could} be obtained with significantly lower network and/or GPU usage if an {\em optimal configuration} of the filtering parameters (\eg video resolution or frame rate) is chosen.
We denote these filtering parameters as \textit{knobs}.
Moreover, the optimal configuration of these knobs can be {\em highly sensitive} to the content of the input data~\cite{awstream,chameleon,casva,accmpeg}. 
For instance, when detecting vehicles in a traffic video, we can significantly reduce the frame rate if the vehicles are moving slowly, and the DNN can still accurately determine the position and movement of the vehicles.

\begin{table}[]
\centering
\footnotesize
\vspace{-0.2cm}
\scalebox{1}{
\begin{tabular}{ccc}
\hline
Knobs                                                                                                                                                                    & \begin{tabular}[c]{@{}c@{}}\Spatial vs.\\ \Temporal\end{tabular} & \begin{tabular}[c]{@{}c@{}}\Uniform vs.\\\Nonuniform\end{tabular} 
\\ \hline
\hline
\begin{tabular}[c]{@{}c@{}}Resolution\\ \cite{awstream,dds,chameleon,vstore,videostorm,deepdecision,adascale,casva}\end{tabular}                                         & \Spatial                                                    & \Uniform                                                                                                                 \\ \hline
\begin{tabular}[c]{@{}c@{}}Quantization parameter\\ (QP) \cite{awstream,vstore,casva}\end{tabular}                                                                       & \Spatial                                                    & \Uniform                                                                                                                            \\ \hline
\begin{tabular}[c]{@{}c@{}}Frame rate\\ \cite{awstream,chameleon,vstore,videostorm,deepdecision}\end{tabular}                                                            & \Temporal                                                    & \Uniform                                                                                                                 \\ \hline
\begin{tabular}[c]{@{}c@{}}Frame filtering thresholds\\ \cite{reducto,glimpse,vigil}\end{tabular}                                                                        & \Temporal                                                    & \Nonuniform                                                                                                                 \\ \hline
\begin{tabular}[c]{@{}c@{}}Region-based QP\\ \cite{accmpeg,dds,eaar,saliency_hotmobile,vcm,neurosurgeon}\end{tabular}                                                    & \Spatial                                                    & \Nonuniform                                                                                                         \\ \hline
\begin{tabular}[c]{@{}c@{}}Audio sampling rate\\ \cite{huang2001spoken, hirsch01_eurospeech, narayanan2018toward, gao2019mixedbandwidth, hokking2016speech}\end{tabular} & \Temporal                                                    & \Uniform                                                                                                                     \\ \hline
\end{tabular}
}
\vspace{0.2cm}
\tightcaption{A list of filtering-related knobs that could be used to reduce bandwidth usage and/or GPU compute without hurting inference accuracy.
} 
\label{tab:rich-set-of-knobs}
\end{table}

\begin{table}[]
\vspace{-0.2cm}
\footnotesize
\setlength\tabcolsep{0.03em}
\centering
\scalebox{1}{
\begin{tabular}{ccccc}
\hline
\multirow{2}{*}{\begin{tabular}[c]{@{}c@{}}{\bf Adaptation}\\ {\bf methods}\end{tabular}}                                       & \multicolumn{4}{c}{\textbf{Knob types}} \\ \cline{2-5} 
 & \begin{tabular}[c]{@{}c@{}} \Uniform\\\spatial \end{tabular} & \begin{tabular}[c]{@{}c@{}} \Nonuniform\\\spatial \end{tabular} & \begin{tabular}[c]{@{}c@{}} \Uniform\\\temporal \end{tabular} & \begin{tabular}[c]{@{}c@{}} \Nonuniform\\ \temporal \end{tabular}  \\ \hline
\hline
\begin{tabular}[c]{@{}c@{}}Profiling\\\cite{chameleon,awstream,videostorm,vstore} \end{tabular}          & \ding{52}                    &                                                                             & \ding{52}                    &                                                        \\ \hline
\begin{tabular}[c]{@{}c@{}}Pixel-filtering\\ heuristics\\\cite{dds,stac,eaar,accmpeg,vigil}\end{tabular}                      &                              & \ding{52}                                                                   &                              &                                                        \\ \hline
\begin{tabular}[c]{@{}c@{}}Frame-filtering \\heuristics~\cite{reducto,glimpse}\end{tabular}                                              &                              &                                                                             &                              & \ding{52}                                              \\ \hline
\begin{tabular}[c]{@{}c@{}}Uniform-filtering \\heuristics~\cite{casva} \end{tabular}          & \ding{52}                    &                                                                             & \ding{52}                    &                                                        \\ \hline
\begin{tabular}[c]{@{}c@{}}\name\\(this work)\end{tabular} & \ding{52}                    & \ding{52}                                                                   & \ding{52}                    & \ding{52}                                              \\ \hline
\end{tabular}
}
\vspace{0.2cm}
\tightcaption{Unlike prior work that works well only for specific knobs, \name can optimize all four types of knobs in Table~\ref{tab:rich-set-of-knobs}.}
\label{tab:knob-categorization}
\end{table}




Depending on how knobs filter the input data, Table~\ref{tab:rich-set-of-knobs} categorizes the knobs along two dimensions. 
\begin{packeditemize}
\item {\em \Spatial vs. \temporal:}
A knob can filter (downsample) input data spatially (\eg lowering video resolution) or temporally (\eg reducing frame rate).
\item {\em \Uniform vs. \nonuniform:}
A knob, \spatial or \temporal, can filter the input data with a coarse or fine granularity.
\end{packeditemize}

Importantly, depending on the type of knob, its optimal configuration can be sensitive to different aspects of the input data. 
For instance, when filtering a video for object detection, a {\em \spatial} and {\em \uniform} knob can lower the resolution of a whole video frame, and its optimal configuration depends on the {\em size} of the objects.
A {\em \spatial} but {\em \nonuniform} knob can lower the resolution only in certain regions in a video frame, and its optimal configuration depends on the {\em location} of the objects. 
Similarly, a {\em \temporal} and {\em \uniform} knob can uniformly reduce the video frame rate to fit {\em how fast} objects move, whereas a {\em \temporal} but {\em \nonuniform} knob can decide whether each frame should be dropped individually to fit {\em when} objects move.


\tightsubsection{Existing adaptation schemes}
\label{subsec:configuration-adaptation}

\begin{figure}
{
\centering
\subfloat[][\normalfont{\textit{Profiling: optimal but too much extra inference.}}\label{fig:profiling}]  
{
    \includegraphics[height=0.38\columnwidth]{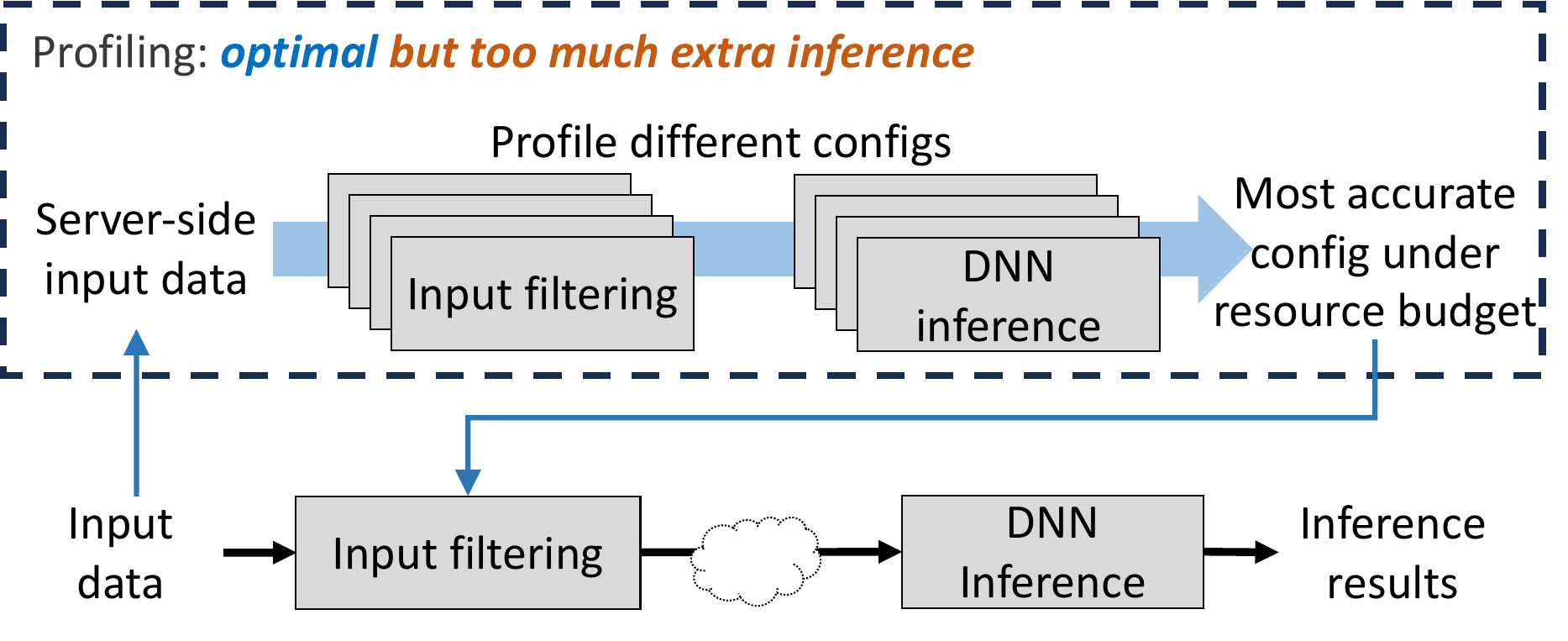} 
    \vspace{-0.3cm}
}
\vspace{-0.3cm}

\subfloat[][\normalfont{\textit{Heuristics: no extra inference but suboptimal.}}\label{fig:heuristics}]  
{
    \includegraphics[height=0.38\columnwidth]{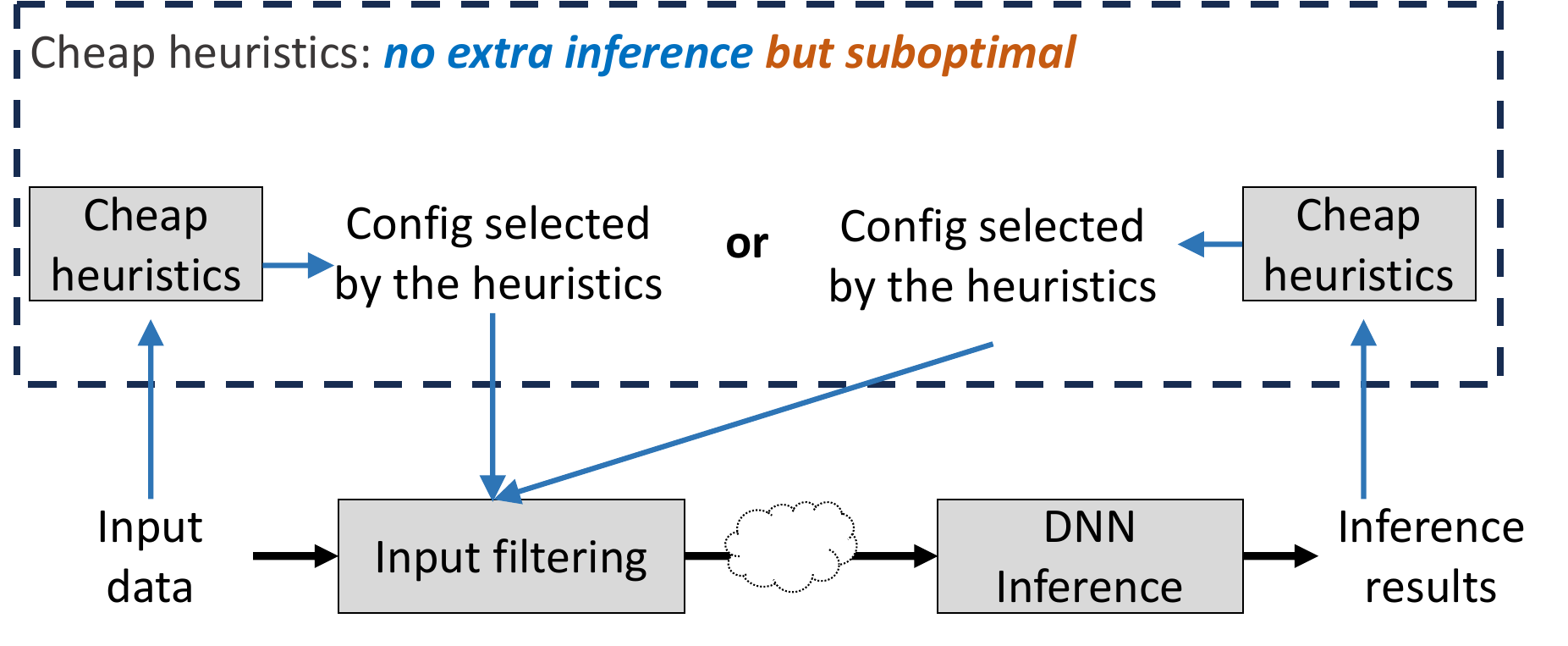} 
    \vspace{-0.5cm}
}
}
\tightcaption{
Illustrating two types of prior work: profiling and heuristics. 
The profiling-based approach can obtain optimal configuration but runs a lot of extra DNN inferences, while heuristics run no extra DNN inference but may pick suboptimal configuration (\S\ref{subsec:configuration-adaptation}).}
\label{fig:prior-work}
\end{figure}

As the content of input data varies over time, these configurations must be adapted timely.
\review{
For instance, it has been shown that increasing the configuration adaptation frequency of frame rate and video resolution from once every 8 seconds to once every 4 seconds increases the percentage of frames with accurate object detection results by 10\%~\cite{chameleon}.
Other works also frequently adapt the configurations of other knobs, such as the encoding quality of each spatial region, at a time scale of a few seconds~\cite{dds,eaar,accmpeg,casva}.
}{This address Reviewer E's comments on the evidence of frequent adaptation}

Therefore, the key research question is to identify if a different configuration is better than the current one. 
There are two high-level categories of techniques.

\myparashort{Profiling-based methods: optimal though slow.} 
The first approach \textit{periodically profiles}
resource usage and
inference accuracy of different configurations by rerunning the same input data using these  configurations, and then picking the best configuration (one that yields high inference accuracy and low resource usage)~\cite{chameleon,awstream,vstore,videostorm}.
Figure~\ref{fig:profiling} illustrates this process.
However, each profiling
needs to run multiple \textit{extra} DNN inferences and thus significantly increases the GPU computation overhead. 
For instance, AWStream~\cite{awstream} runs \textit{full} profiling on just 3 knobs (resolution, frame rate, and quantization parameter) on a \textit{10-second}
video, and its GPU computation is equivalent to running normal DNN inference on  
more than an \textit{8-minute} video (\ie an almost 50$\times$ increase in GPU usage). 
As a result, if GPU resource is limited,
this approach must either profile less frequently, causing it to use outdated configurations when the optimal configuration changes
(as shown in Figure~\ref{fig:adaptation-behavior}), or profile only fewer configurations, which may miss the optimal configuration.
\S\ref{sec:eval} will also test smarter variants of profiling (profiling only top $k$ configurations and profiling fewer values per knob) and shows that they still use much GPU resource for extra inference and are thus slow to adapt to change in input data.

\mypara{No extra inference though suboptimal}
An alternative approach~\cite{reducto,vigil,glimpse,accmpeg,casva,filterforward,dds,eaar,elf,fischer2021saliency,stac} avoids extra inference and instead selects new configurations by analyzing either the input data or the DNN intermediate inference results and filtering out those data that are not of the interest of the final DNN using simple heuristics.
We illustrate this process in Figure~\ref{fig:heuristics}.
This approach can adapt frequently as it does not require extra inference by the final DNN.
However, they may select a suboptimal configuration.
Here, we discuss three types of heuristics. 


\begin{packeditemize}

\item \textit{Heuristics adapting \nonuniform \temporal knobs:} 
One line of work~\cite{reducto,glimpse,eaar} selects the frames most worthy for analysis by applying \nonuniform \temporal knobs (\eg by computing the difference between the current frame and previously analyzed frame
and sends the current frame out for DNN inference if the difference is greater than a threshold).
As elaborated in \S\ref{sec:intro}, this method fails when the background pixels in a video change a lot between frames, making high frame difference a poor signal to decide if a frame should be analyzed or not.

\item \textit{Heuristics adapting \nonuniform \spatial knobs:}
Another line of work~\cite{dds,eaar,vigil,accmpeg,fischer2021saliency} reduces bandwidth usage by calculating the regions of interest, and assigns these regions with higher encoding quality.
Some of these work identifies regions of interest by running a \textit{shallow neural network} on each frame (\eg MobileNet-SSD~\cite{accmpeg,ssd}).
However, as acknowledged in one of the latest work~\cite{accmpeg}, such shallow neural network rarely recognizes small vehicles and thus fails to encode them in high quality.
Other heuristics try to leverage information naturally emitted from the final DNN, such as region proposals (an intermediate output of some DNNs indicating which regions in the input data may contain objects).
This approach can also adapt quickly, but as shown in~\cite{accmpeg}, region proposals only indicate whether {\em some} objects are in the region, rather than whether raising the quality of that region will likely affect the ability for the DNN to detect the objects of interest.

\item \textit{Heuristics adapting \uniform knobs}:
Recent proposals also propose heuristics based on reinforcement learning~\cite{casva} to adapt \uniform knobs such as resolution and frame rate. This heuristic relies on the file size of the encoded video to adapt the knobs.
However, the file size of the encoded video cannot indicate whether the object of interest appears in the video or not, making this approach unable to timely raise the frame rate and resolution when the object of interest appears.

\end{packeditemize}


\reviewreview{
\mypara{Bayesian Optimization}
Though less explored in the literature of streaming media analytics systems, an alternative method can treat the analytic system as a black box and use Bayesian Optimization (BO) to search for the best configuration~\cite{cherrypick,ganet}.
In contrast, gradient-based optimization, which our work belongs to, has been shown to converge faster than BO  in a discrete configuration space with a concave objective function.
In theory, after evaluating $k$ configurations, gradient-based optimization's function value has a gap-to-optimal of $O(\frac{1}{k^2})$~\cite{schmidt2011convergence,nesterov2013gradient} whereas BO's gap-to-optimal is $O(\frac{1}{\sqrt{k}})$~\cite{kawaguchi2015bayesian,de2012exponential}.
As a result, gradient-based optimization, rather than BO, has been widely used in various discrete spaces with concave functions (\eg ~\cite{ramazanov2011stability,chistyakov2015stability,zhang2022gradient,chizat2021convergence}).

In the case of streaming media analytics, the configuration values are discrete and, as shown later in \S\ref{subsec:limitation}, the objective function between the configuration and accuracy is mostly concave. 
Thus, we choose to use gradient-based optimization, rather than BO, and utilize DNN's differentiability to estimate the (numerical) gradients without extra DNN inference.

}{This addresses concerns regarding Bayesian Optimization}

\mypara{Designed for specific configurations or DNNs}
Moreover, most techniques are studied and evaluated {\em only} on RGB video feeds, and it is unclear whether they will be effective on other types of data (\eg LiDAR point cloud, depth map, or audio waves).
Even in video analytics, the heuristics often work well only for {\em specific knobs} (we show the applicability of prior works on different types of knobs in Table~\ref{tab:knob-categorization}).
For example, Reducto~\cite{reducto}'s frame-different detector is ill-suited to tune the encoding quality parameter of each frame, which depends on the content of the new frame, rather than how it differs from the previous frame.
Likewise, DDS~\cite{dds} tunes the encoding quality parameters of each fine-grained spatial block to improve inference accuracy on each frame, but it is not aware of whether the current frame is redundant given past inference results, thus it cannot generalize to knobs such as frame selection.
Also, as the amount of extra inference scales with the number of knobs, profiling-based approach~\cite{awstream,chameleon,vstore,videostorm} is suboptimal when handling more than 10 knobs (which is the case for  \nonuniform knobs).

\mypara{Summary} 
In short, previous work fails to meet at least one of the three requirements: {\em (i)} adapt timely, {\em (ii)} converges to a near-optimal configuration, and {\em (iii)} apply to a wide range of configurations and applications.

\tightsection{Design of \name}
\label{sec:design}

\begin{figure}
    {
    \centering
    {
        \includegraphics[width=.99\columnwidth]{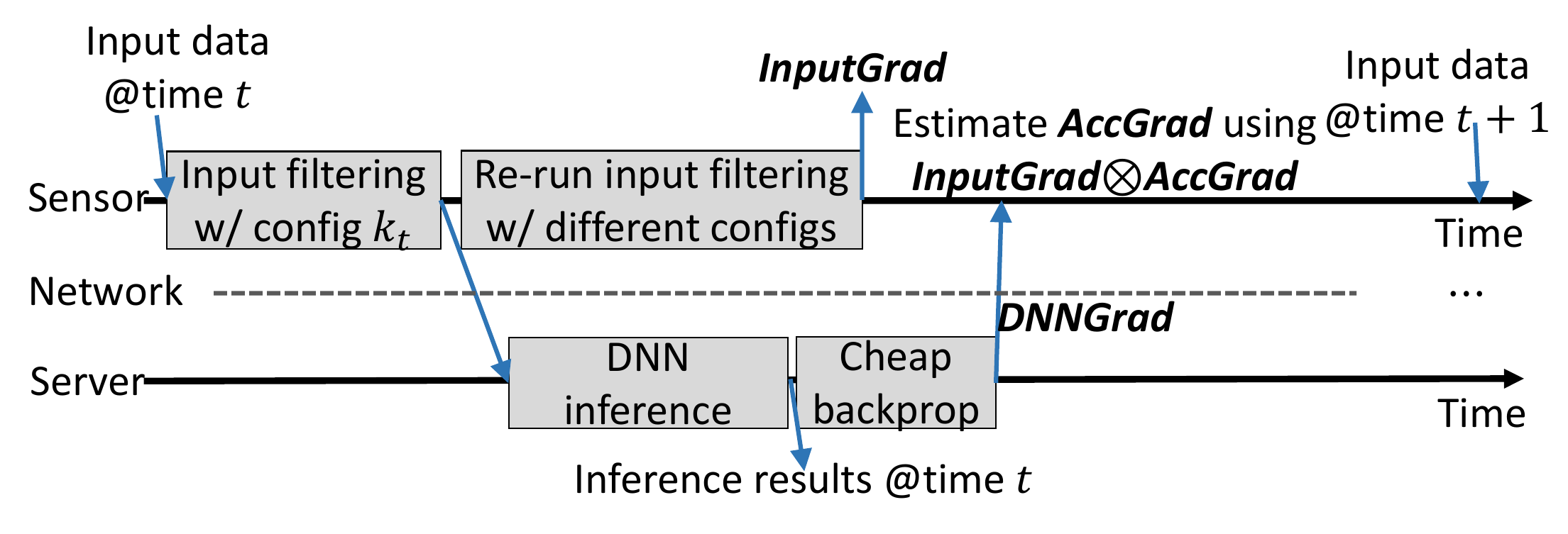}
        
    }
    }
    \vspace{-0.2cm}
    \tightcaption{Illustrating how \name estimates \accgrad using \inputgrad and \dnngrad using the sensor and the server.}
    \label{fig:workflow}
\end{figure}



We present \name, a configuration adaptation system that aims at improving along the three requirements mentioned above --- frequent, near-optimal adaptation on various configuration knobs. 
The basic idea of \name is to frequently estimate 
the gradient of how inference accuracy changes with each configuration knob, which we refer to as \accgrad.
\name then performs gradient ascent based on \accgrad to update the configuration, with an objective metric that combines inference accuracy and resource usage (\S\ref{subsec:adaptation-goal}). 
To efficiently estimate \accgrad, \name approximates it by estimating \outputgrad, another metric that can be efficiently calculated by leveraging DNN's inherent differentiability (\S\ref{subsec:outputgrad}) and statistically correlated with \accgrad (\S\ref{subsec:outputgrad-properties}).
Finally, we will discuss the caveats of \name's gradient-based adaptation (\S\ref{subsec:limitation}).

\tightsubsection{Terminology}
\label{subsec:terminology}


To adapt configuration for various DNN pipelines
, we introduce a unified terminology to describe their input and output.
Table~\ref{tab:notations} summarizes the notations and their meanings.

We split the input data into fixed-length (by default, one-second-long) intervals, called {\em adaptation intervals}.
In the $t^{\textrm{th}}$ adaptation interval, the input data $\Data_t$ is first preprocessed with the current configuration $\Config_t$ (\eg video resolution and frame rate) to get the DNN input $\Input(\Config_t; \Data_t)$, 
and the DNN takes it as input and returns $Res(\Config_t; \Data_t)$ as output.
DNN output $Res(\Config_t, \Data_t)$ contains multiple {\em elements}. 
For the DNNs in our considered tasks, each element is associated with a confidence {\em score}, and only elements with a score over a confidence {\em threshold} $\theta$ will be counted towards accuracy. 

The definitions of an element $e$ and input data $\Data$ vary with the considered application.
An element is a detected object in object detection, a detected text in audio-to-text translation, or a segmented mask of a detected instance in instance segmentation.
Input data can be a sequence of RGB frames in videos, point-cloud frames in LiDAR, or an audio wave segment in audio data.


Based on these notations, we define:
\begin{packeditemize}
\item $Acc(\Config_t; \Data_t)$ is the accuracy of the inference results generated using input data $\Data_t$ and configuration $\Config_t$. 
Following prior work~\cite{awstream,chameleon,dds,accmpeg,ekya,boggart}, we define accuracy as the \textit{\textbf{similarity}} between the current inference results and the inference results generated from the most resource-consuming configuration $\Config^*$.\footnote{This notion of accuracy is well studied in the literature. While it does not compare DNN output with human-annotated ground truth, it better captures the impact of adapting configurations on inference results.} 
\item $\Output(\Config_t; \Data_t)$ is the \summary of the inference results generated using input data $\Data_t$ and configuration $\Config_t$. We define \summary as the number of elements with confidence scores above the confidence threshold (see \S\ref{subsec:outputgrad}).
\item $\Resource(\Config_t)$ is the resource usage (bandwidth and/or GPU cycles, defined per application) at the $t^{\textrm{th}}$ adaptation interval. 
\end{packeditemize}

\review{
We may omit the time label $t$ and the input data $\Data_t$ for simplicity when the corresponding variables are of the same interval as the current input data $\Data_t$.}{This addresses Reviewer A's concern on the omission of time label}


\tightsubsection{Adaptation goal and gradient-ascent}
\label{subsec:adaptation-goal}

The adaptation goal of \name is to pick the configurations for all adaptation intervals such that the following weighted sum between the accuracy $Acc$ and the resource usage $\Resource$ is maximized:

\vspace{-0.2cm}
\begin{equation}
\sum_{t=1}^T
\underbrace{Acc(\Config_t)}_{\text{\sf \scriptsize \textcolor{black!85}{accuracy of $t^{th}$ interval}}}
- \lambda \underbrace{\Resource(\Config_t)}_{{\text{\sf \scriptsize \textcolor{black!85}{resource usage of $t^{th}$ interval}}}},
\label{eq:objective}
\end{equation}

\noindent where
$\lambda$ is a hyperparameter that governs the tradeoff between accuracy and resource usage: higher $\lambda$ will emphasize resource-saving and lower $\lambda$ will emphasize accuracy improvement.
We will vary $\lambda$ and examine its effect on \name in \S\ref{sec:eval}.
Note that this objective can be extended to perform {\em budgeted} optimization (\eg to maximize accuracy under a budget of resource usage, we can add a change the term $\Resource(\Config_t)$ to a large penalty when resource usage extends the budget).
We leave this extension to future work.

To optimize towards this goal, \name uses an online \textbf{\textit{gradient-ascent strategy}} (illustrated in Figure~\ref{fig:workflow}).
Concretely, \name derives a new configuration $\Config_{t+1}$ based on the current configuration $\Config_t$ using the following Equation:

\vspace{-0.3cm}
\begin{@empty}
\footnotesize
\begin{align}
\Knob_{t+1,i}=\Knob_{t,i}+ \alpha \left(
\underbrace{
\frac{\Acc(\Config_t+\Delta \Knob_i) - \Acc(\Config_t) }{\Delta \Knob_i}
}_{\text{\sf \scriptsize \textcolor{black!85}{grad. of accuracy (\textbf{\accgrad})}}}
- 
\lambda 
\underbrace{\frac{\Resource(\Config_t+\Delta \Knob_i) -\Resource(\Config_t) }{\Delta \Knob_i}
}_{\text{\sf \scriptsize \textcolor{black!85}{grad. of resource usage}}}
\right),
\label{eq:ideal-adaptation}
\end{align}
\end{@empty}
\vspace{-0.1cm}

\noindent where $\Knob_{t+1,i}$ is the configuration of the $i^{th}$ knob in the next adaptation interval, $\Delta\Knob_i$ is a small increase on the $i^{th}$ knob, $\alpha$ is the learning rate.
We will discuss the convergence of this gradient-ascent strategy and extend it to discrete configuration values in \S\ref{subsec:limitation}.
For now, we assume the configuration of each knob can be tuned continuously. 
Also, while more advanced gradient-based optimization (\eg Nesterov's Accelerated Gradient Descent~\cite{nesterov2013gradient}) exists, \name chooses a standard gradient-ascent strategy to make sure the source of improvement is from the gradient itself instead of advanced optimization techniques.


The gradient-ascent strategy in Equation~\ref{eq:ideal-adaptation}
requires calculating the gradient of accuracy (hereinafter \accgrad) and the gradient of resource usage along each knob $i$.
We calculate the gradient of resource usage by using its definition (\ie we directly calculate the bandwidth and GPU computation consumption of configuration  $\Config_t+\Delta \Knob_i$ and then calculate$ \frac{\Resource(\Config_t+\Delta \Knob_i) -\Resource(\Config_t) }{\Delta \Knob_i}$)\footnote{Note the calculating the resource usage of a configuration does not require DNN inference, as calculating the bandwidth usage requires no inference, and the GPU computation largely depends on the shape of DNN input rather than the content of the input.}. 

\begin{figure}[t]
    \centering
    \subfloat[][Calculating \outputgrad using its definition]
    {
        \includegraphics[width=0.99\columnwidth]{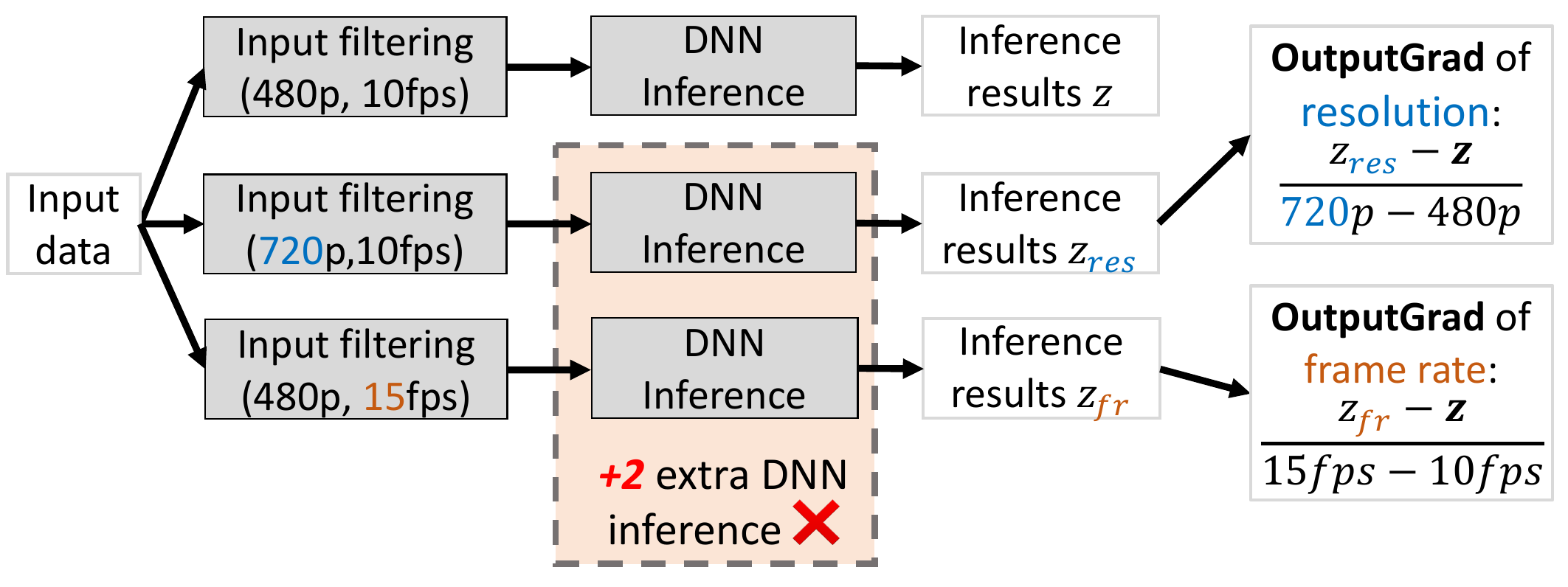} 
        \label{subfig:gradient-before}
    }
    \\
    \subfloat[][Calculating \outputgrad using \name]
    {
        \includegraphics[width=0.99\columnwidth]{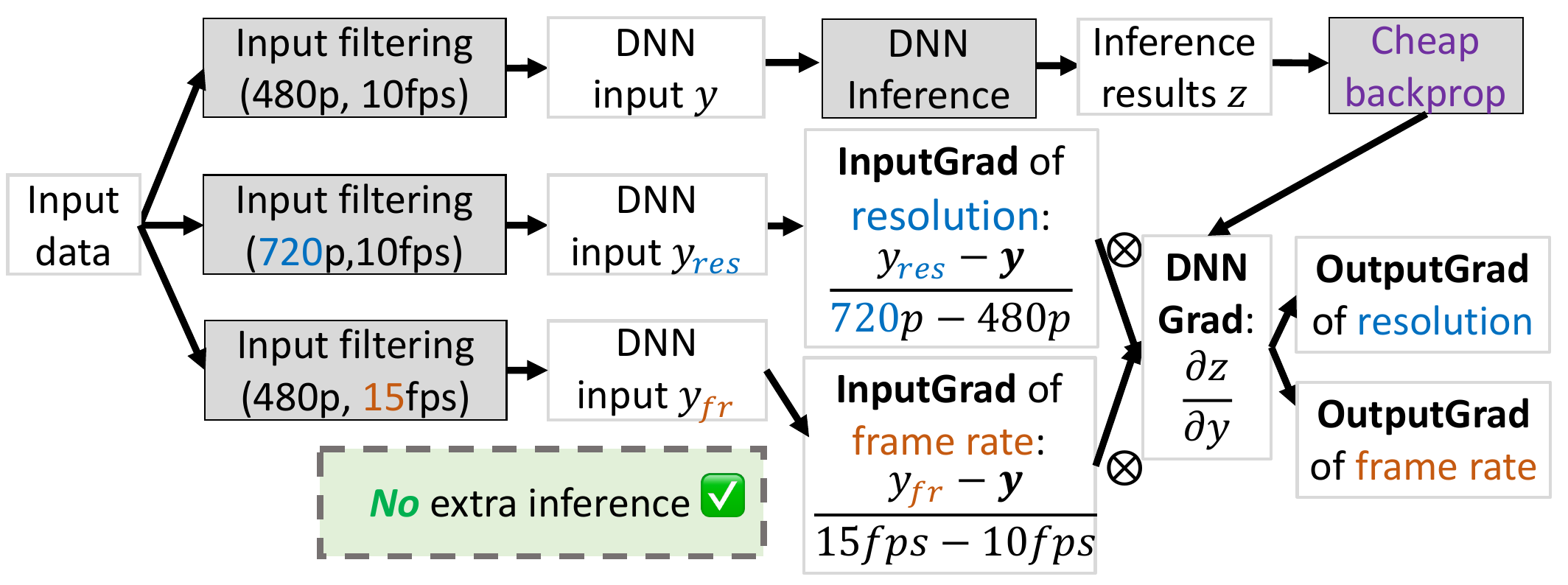} 
        \label{subfig:gradient-after}
    }
    \tightcaption{Comparison between calculating \outputgrad naively and calculating \outputgrad using \name. \name calculates \outputgrad with \textbf{no extra inference}.}
    \label{fig:gradient-contrast}
\end{figure}

However, if we calculate \accgrad by its definition, it would be prohibitively expensive (see Figure~\ref{subfig:gradient-before}). 
Obtaining $Acc(\Config)$ of any configuration $\Config$ will require running inference on the most resource-intensive configuration $\Config^*$.
Moreover, to get the \accgrad of each knob would require running an extra inference to test the impact of a small change in configuration on DNN output.

\tightsubsection{Fast approximation of \accgrad}
\label{subsec:outputgrad}

To estimate \accgrad efficiently, we introduce a new metric, called \outputgrad, to approximate it.
Here, we define \outputgrad and explain how to calculate it efficiently. 
The next subsection will show its statistical correlation with \accgrad.

\mypara{DNN \summary}
Recall from \S\ref{subsec:terminology} that the DNN output contains multiple {\em elements}, each associated with a confidence {\em score}.
An element can be a detected object in object detection or a detected text in audio-text translation.
Since only elements with a score over the confidence {\em threshold} will be counted towards accuracy, we define {\em \summary} of a DNN output by the number of elements in an adaptation interval whose confidence scores exceed the confidence threshold (see Equation~\ref{eq:outputgrad} for a formal definition).\footnote{This definition applies to all applications mentioned in Table~\ref{tab:pipeline}, but future work is needed to extend the definition to a wider range of applications.}
We use $\Output(\Config_t)$ to denote the \summary 
of DNN output $Res(\Config_t; \Data_t)$ under configuration $\Config_t$. 

\Summary offers a single-value summarization of the DNN output whose change indicates the change in inference accuracy
(More rationale of \summary in \S\ref{subsec:outputgrad-properties}).

\mypara{\outputgrad definition}
\outputgrad is defined as how much a small change in the current value of each knob changes the \summary.
The \outputgrad of knob $\Knob_i$ can be written as
$$
\frac{\Delta \Output}{\Delta k_i} = \frac{\Output{(\dots,\Knob_{t,i}+\Delta \Knob_i,\dots)}-\Output{(\dots,\Knob_{t,i},\dots)}}{(\Knob_{t,i}+\Delta \Knob_i) - \Knob_{t,i}}.
$$


To help understand \outputgrad, we use a simple 
video analytics system that feeds each video frame to a vehicle detection DNN, using a confidence score of 0.5. 
Figure~\ref{subfig:gradient-before} illustrates \outputgrad in this example.
Following prior work~\cite{chameleon,vstore,videostorm,casva}, we assume the system can tune two configuration knobs, frame rate, and resolution.
Suppose that the current configuration is 10 frames per second (fps) and 480p.
When taking 1-second input, the DNN outputs 34 vehicle bounding boxes with confidence scores greater than 0.5 on these 10 frames, an average of 3.4 vehicles per frame.
If we slightly increase the frame rate from 10fps to 15fps, the DNN outputs 66 vehicle bounding boxes with confidence scores greater than 0.5, on average 4.4 vehicles per frame.
Then the \outputgrad of the current frame rate is 
\begin{@empty}
\footnotesize
$$
\frac{\Delta \Output}{\Delta \text{framerate}}=\frac{\Output{(480p,15fps)}-\Output{(480p,10fps)}}{15fps-10fps}=\frac{4.4-3.4}{15-10}=0.2.
$$
\end{@empty}
Similarly, the \outputgrad of the resolution is:
\begin{@empty}
\footnotesize
$$
\frac{\Delta \Output}{\Delta \text{resolution}}=\frac{\Output{(720p,10fps)}-\Output{(480p,10fps)}}{720p-480p}=\frac{5.8-3.4}{720-480}=0.01.
$$
\end{@empty}




\mypara{How to calculate \outputgrad efficiently} 
The insight of \name is that \outputgrad can be computed efficiently \textit{with no extra DNN inference} by taking advantage of the \textit{\textbf{differentiability of DNNs}}. 
As our considered application pipelines affect the inference results by altering the DNN input through knobs,
the \outputgrad of a knob can be written as the inner product of two separate gradients (illustrated in Figure~\ref{fig:workflow}), each can be computed efficiently:
\begin{packedenumerate}
\item {\em \dnngrad:} How the DNN's output changes with respect to the DNN's input. 
\item {\em \inputgrad:} How the DNN's input changes with respect to the knob's configuration. 
\end{packedenumerate}

Formally, the \outputgrad of $\Knob_i$ on \Data can be expressed by:
\begin{@empty}

$$
{\color{black}\underbrace{\highlight{white}{$\frac{\displaystyle \Delta \Output}{\displaystyle \Delta \Knob_i}$}}_{\text{\sf \scriptsize \textcolor{black!85}{\outputgrad}}}}=
{\color{black}\underbrace{\highlight{white}{$\frac{\displaystyle \Delta \Output}{\displaystyle \Delta \Input(\Config)}$}}_{\text{\sf \scriptsize \textcolor{black!85}{\dnngrad}}}}\otimes
{\color{black}\underbrace{\highlight{white}{$\frac{\displaystyle \Delta \Input(\Config)}{\displaystyle \Delta \Knob_i}$}}_{\text{\sf \scriptsize \textcolor{black!85}{\inputgrad}}}},$$
\end{@empty}
where $\otimes$ means inner product, and $\Input(\Config)$ is the DNN input of configuration $\Config$.

The decoupling makes the calculation of \outputgrad much more resource-efficient for three reasons:
\begin{packeditemize}
\item \dnngrad can be estimated by running one backpropagation
.
This is a constant GPU-time operation without extra inference as it reuses the layer-wise features computed as part of the DNN inference of the current configuration.
\S\ref{sec:cost} will further reduce the overhead of backpropagation.

\item \dnngrad only needs to be calculated once and can be re-used to derive \outputgrad of different knobs without extra DNN inference.
This is because the \dnngrad describes DNN's sensitivity to its input and thus remains largely similar if the change in DNN input is not dramatic.
For instance, the \dnngrad of an object-detection DNN is high on pixels associated with key visual features of an object.

\item Finally, computing \inputgrad does not require running the final DNN.

\end{packeditemize}



To reuse the same example from Figure~\ref{subfig:gradient-before}, Figure~\ref{subfig:gradient-after} shows how \outputgrad is calculated from \dnngrad and \inputgrad:

\vspace{-0.3cm}
\begin{@empty}
\scriptsize
\begin{align*}
\frac{\Output_{(480p,15fps)}-\Output_{(480p,10fps)}}{15fps-10fps} &\approx \frac{\Input_{(480p,15fps)}-\Input_{(480p,10fps)}}{15fps-10fps} \otimes \frac{\Delta \Output}{\Delta \Input}\bigg|_{\Input=\Input_{(480p,10fps)}},\\
\frac{\Output_{(720p,10fps)}-\Output_{(480p,10fps)}}{720p-480p} &\approx \frac{\Input_{(720p,10fps)}-\Input_{(480p,10fps)}}{720p-480p} \otimes \frac{\Delta \Output}{\Delta \Input}\bigg|_{\Input=\Input_{(480p,10fps)}},
\end{align*}
\end{@empty}
where $\otimes$ represents inner product and $\frac{\Delta \Output}{\Delta \Input}|_{\Input=\Input_{(480p,10fps)}}$ is \dnngrad of the current configuration $(480p,10fps)$.
Figure~\ref{subfig:gradient-after} shows that \name saves extra GPU inference by running backpropagation once and re-using the backpropagation results for different knobs.
This example has two knobs, thus the saving may seem marginal, but our evaluation shows more savings when \name is used to adapt more knobs in more realistic DNN analytics systems (Figure~\ref{fig:fine-grained-knobs}).

\subsection{Relationship between \outputgrad and \accgrad}
\label{subsec:outputgrad-properties}

The definitions of \outputgrad and \accgrad differ, yet they are closely related both theoretically and empirically
This can be intuitively explained as follows: a high \outputgrad means a great change in the inference output, which often causes a greater change in accuracy and thus a high \accgrad.

\mypara{Theoretical correlation}
To formalize this correlation between \outputgrad and \accgrad, we prove that they are statistically correlated under specific \textbf{\textit{assumptions}}. 
\begin{packeditemize}
\item First, we define the accuracy of DNN output as the number of correctly-identified elements (including both true positive and true negative), minus the number of wrongfully-identified elements (in our proof we use a differentiable approximation of this accuracy function, see Equation~\ref{eq:accuracy-definition} for formal definition).
Acknowledging that this definition of accuracy differs from the metric that is widely used in prior work (\eg F1 score~\cite{dds,eaar,awstream,chameleon,accmpeg,casva,vstore,ekya,videostorm}), we observe that, if a configuration has higher accuracy than another configuration under one accuracy metric, it is likely that this configuration also has higher accuracy under other accuracy metrics.
\item Second, the accuracy of DNN output increases when a knob changes towards using more resources (\eg higher resolution or selecting more frames). This observation aligns with prior work~\cite{awstream,chameleon,dds,eaar,accmpeg,vstore,videostorm,reducto,boggart}.
\item 
Third, different elements do not overlap (\eg object bounding boxes do not mutually overlap, or the detected words in the audio do not overlap).
This assumption holds as a wide range of DNNs perform non-maximum suppression~\cite{nms,detectron2,seqnms} to remove overlapped elements.
\end{packeditemize}
Formally, if these assumptions hold, we can formally prove that 
\begin{@empty}

$$
\underbrace{\frac{\partial Acc(\Config)}{\partial \Config}}_{\text{\sf \footnotesize \textcolor{black!85}{\accgrad}}} = \bigg| \underbrace{\frac{\partial \Output(\Config)}{\partial \Config}}_{{\text{\sf \footnotesize \textcolor{black!85}{\outputgrad}}}}\bigg|
$$
\end{@empty}
.
The proof can be found in \S\ref{sec:proof}.




While there can be exceptions to the conditions required by our proof, these conditions corroborate the observations in previous studies.
For instance, higher encoding resolution leads to higher bandwidth usage and likely more accurate inference, or higher frame rate leads to higher GPU usage and likely more accurate results~\cite{awstream,chameleon,dds,eaar,accmpeg,vstore,videostorm,reducto,boggart}.





\begin{figure}
    \centering
    \includegraphics[width=0.7\columnwidth]{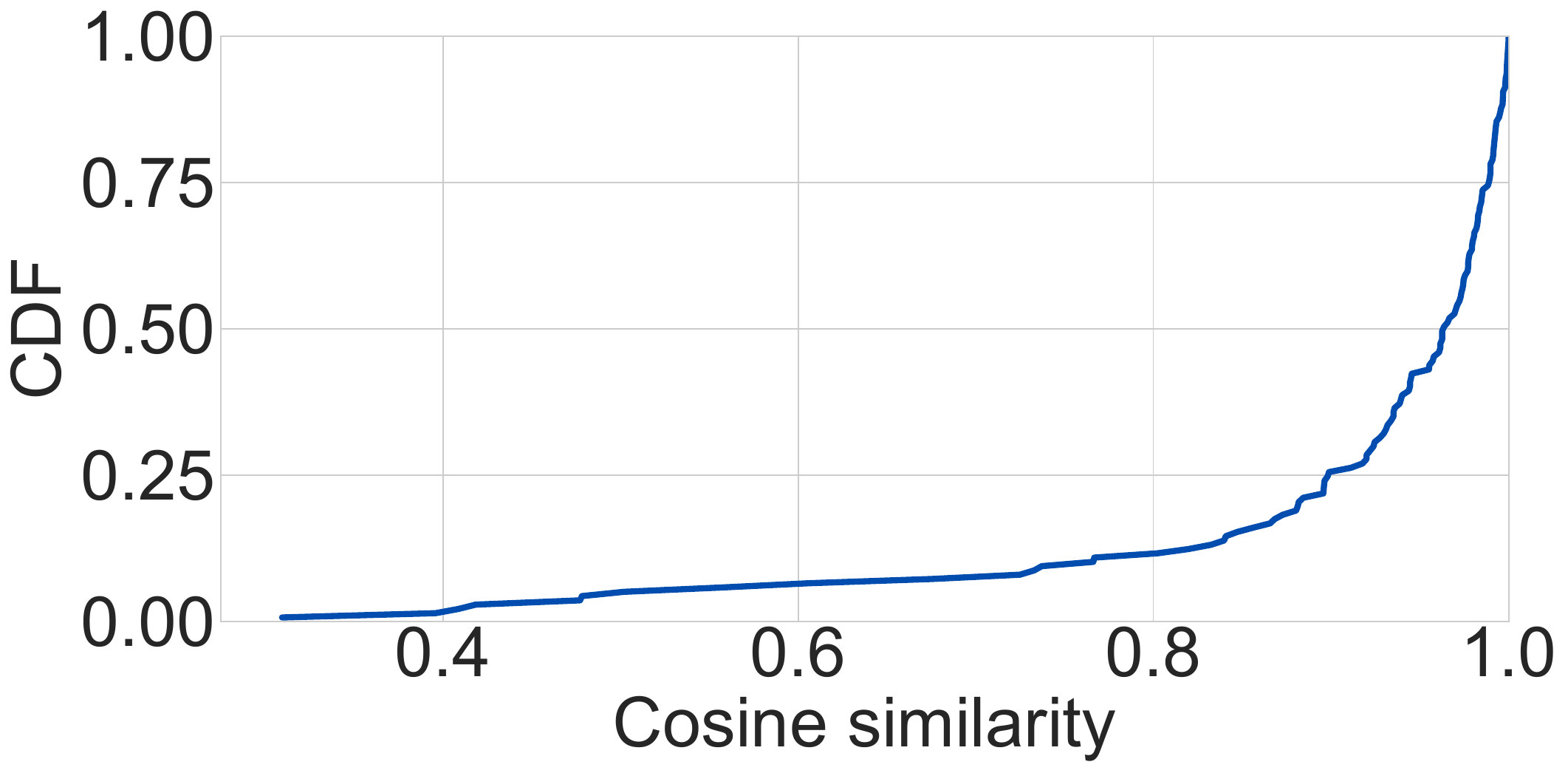}
    \tightcaption{The CDF of the cosine similarity between \accgrad and \outputgrad across different configurations and videos. The average cosine similarity is over 0.91.}
    \label{fig:accgrad-outputgrad-cosine-similarity}
\end{figure}

\mypara{Empirical correlation}
\review{
We empirically validate the correlation between \accgrad and \outputgrad using a streaming media analytics pipeline (pipeline~\circled{a}) from the {\sf{Downtown}} dataset (refer to \S\ref{sec:eval} for details).
For each configuration in pipeline~\circled{a}, we derive \accgrad from its definition and \outputgrad through backpropagation on the first 10 seconds of each video in {\sf{Downtown}} dataset.
Since both \accgrad and \outputgrad are vectors (with each element being the gradient for a knob), we measure their correlation by cosine similarity.
Figure~\ref{fig:accgrad-outputgrad-cosine-similarity} displays the CDF of cosine similarity across different configurations and videos. The average cosine similarity exceeds 0.91, indicating a strong correlation between \accgrad and \outputgrad.
}
{This addresses Reviewer A's concern about how well \outputgrad approximates \accgrad}

\tightsubsection{Caveats and benefits}
\label{subsec:limitation}

While our evaluation shows \name's gradient-ascent strategy performs well for a wide range of DNN tasks, input data types, and configuration knobs, it is important to understand its limits and why it works in practice.

\noindent \textbf{Convergence of \name on changing input data:}
\review{
\name takes a gradient-ascent strategy to adapt configurations, where \name derives a new configuration based on the input data of the current adaptation interval and applies it to the next interval. 
This approach will not converge if the input changes dramatically between intervals.
That said, the input data in our benchmark (and other papers~\cite{awstream,chameleon}) incurs drastic change on the scale of tens of seconds (\eg when driving in the countryside, the vehicle appears on a scale of tens of seconds), 
whereas \name empirically converges to a near-optimal configuration within 3-5 intervals (seconds) as shown in Figure~\ref{fig:converge-fast}.
Thus, as long as the best configuration lasts for at least a few seconds, \name can identify the best configuration faster than profiling-based methods.



The reason behind the convergence of \name is that the relationship between configuration $\Config$ and its corresponding accuracy $\Acc(\Config)$ 
is roughly \textbf{\textit{concave}}, allowing the gradient-ascent strategy of \name to converge to a near-optimal configuration.
This concavity also appears in prior work~\cite{dds,accmpeg,awstream,eaar}.
This concavity means that as configuration changes in the direction of using more resources, the improvement in accuracy will diminish as accuracy approaches 100\%.
For example, as a video analytic system increases the video encoding resolution, the pixel change becomes more marginal (mostly on ``high-frequency'' details), causing the DNN output to stabilize (objects' confidence scores will change more slowly and less frequently cross the confidence threshold, leading to fewer newly detected objects) and leading to less improvement on accuracy~\cite{awstream,chameleon,vstore,videostorm,casva}.
A well-known consequence of such concavity is that gradient-ascent can eventually converge to a global optimal configuration~\cite{gradient-descent-convex,flaxman2004online}. 

}{This address Reviewer A's concern on the convergence of \name.}

\mypara{Why \name beats alternatives}
\name outperforms previous work on all three requirements (\ie adapt frequently, converge to a near-optimal configuration, and generalize to different types of knobs and analytic tasks).

First, compared to profiling-based work, \name adapts much more frequently (as \name updates its configuration \textit{at every second} but profiling-based work updates its configuration once every minute~\cite{awstream}) and earlier (as \name adapts to the change of input content right at the next second but the profiling-based work need to wait till the next profiling).

Second, compared to heuristics-based methods, \name's gradient ascent can converge to a {\em closer-to-optimal} configuration.
Most heuristics used in prior work adapt configuration by analyzing only the input data, rather than how the data might affect the final DNN's output and accuracy.
By contrast, \accgrad directly indicates that DNN accuracy varies with a small change in the configuration. 

Third, we show that \name can generalize to 4 analytic tasks (vehicle detection, human detection segmentation, audio-to-text), 5 types of input data (\eg LiDAR videos, audio waves), and 5 types of knobs (\eg video codec knobs, frame filtering thresholds) in our evaluation (\S\ref{sec:eval}). 
That said, we have not tested if \name will work for applications outside streaming media analytics (\eg text generation and image generation) and knobs that alter the final DNN itself (\eg DNN selection).


\review{We want to clarify that similar to prior work~\cite{awstream,chameleon,casva}, \name leverages idle time to perform adaptation. Thus, \name does not delay the generation of inference results.}{This addresses Reviewer C's concern on the delay of \name}

\mypara{Running gradient ascent over discrete configuration space}
Before running \name, we first linearly normalize all knob values to [0,1] and make sure that the increase in knob value corresponds to the increase in resource usage.
We then calculate the updated knob value (Equation~\ref{eq:ideal-adaptation}) and configure each knob using the configuration value closest to the updated value.
We want to clarify that running gradient ascent on top of discrete configuration space can converge to a near-optimal configuration when the objective function (\ie the weighted sum between accuracy and resource usage) is concave~\cite{ramazanov2011stability,chistyakov2015stability}.

\tightsection{Optimization of \name}
\label{sec:cost}

Though \outputgrad estimation in \name does not involve extra DNN inference, it imposes three types of overhead:
\begin{packedenumerate}
    \item {\em GPU overhead} to run backpropagation and obtain \dnngrad.
    \item {\em CPU overhead} to filter input data multiple times and compute \inputgrad of each knob.
    \item {\em Bandwidth overhead} to stream the \dnngrad from the DNN back to the sensor (as shown in Figure~\ref{fig:workflow}). 
\end{packedenumerate}
The first two overheads are present in both the distributed setting (where the sensor streams DNN input through a bandwidth-constraint link to a server for DNN inference) and non-distributed settings (where the sensor and DNN co-locate).
The third overhead is specific to the distributed setting. 
This section presents the optimizations used by \name to 
reduce each overhead.

We want to clarify that similar to prior work~\cite{awstream,chameleon,casva}, \name leverages idle time to adapt. Thus, these overheads of \name do not delay the generation of inference results.

\tightsubsection{Reducing GPU computation overhead}
\label{subsec:saliency_sampling}

\name uses two techniques to reduce the GPU overhead of backpropagation.


\mypara{Removing unneeded computation} 
The standard implementation of DNN backpropagation is designed for DNN training, which computes not only gradients with respect to the DNN input (\ie \dnngrad) but also gradients with respect to DNN weights.
While the gradients with respect to DNN weights are crucial for DNN training (in order to update DNN weights), they are unnecessary to compute \dnngrad.
We note that this optimization does {\em not} change \dnngrad.

\begin{figure}[t]
    \centering
    \includegraphics[width=0.7\columnwidth]{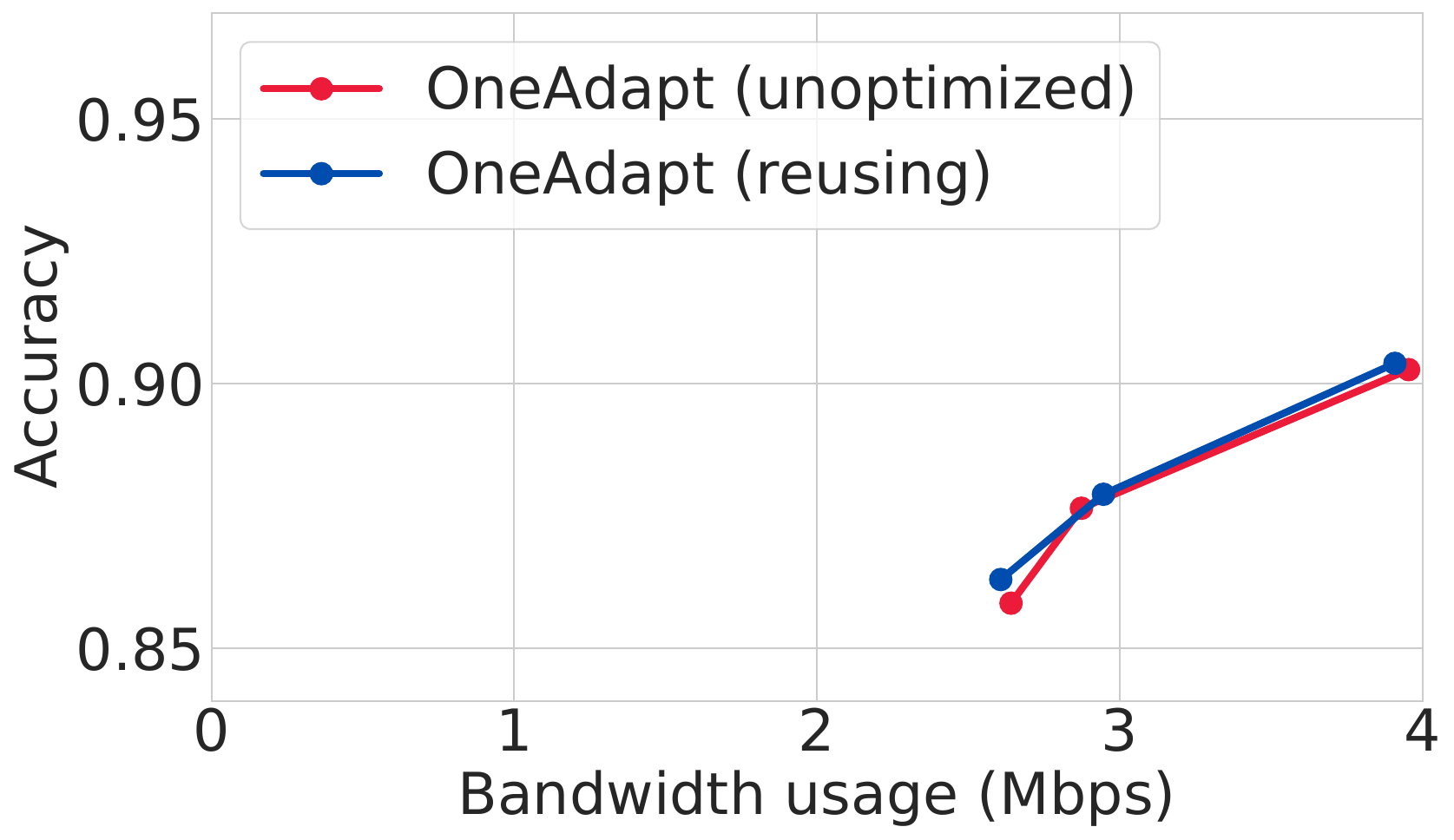}
    
    \tightcaption{Benchmarking the impact of \dnngrad reusing on pipeline~\circled{a}. \dnngrad reusing does not reduce the accuracy or increase the resource usage of \name.
    }
    \vspace{-0.2cm}
    \label{fig:reduce-overhead-min-impact}
\end{figure}

\mypara{\dnngrad reusing} 
To further reduce GPU computation overhead, we run backpropagation to compute \dnngrad on the last frame or segment of the current adaptation interval, and reuse this \dnngrad on other data frames in the current adaptation interval.
In the context of video analytics, it can be intuitively understood that although the exact value of \dnngrad may vary across different frames, the \textit{spatial} areas that have high \dnngrad tend to be stable within several consecutive frames.
Though \dnngrad reusing alters \dnngrad, Figure~\ref{fig:reduce-overhead-min-impact} shows that this optimization has little impact on the resource--accuracy trade-off of \name.

\begin{figure}
    \includegraphics[width=0.99\columnwidth]{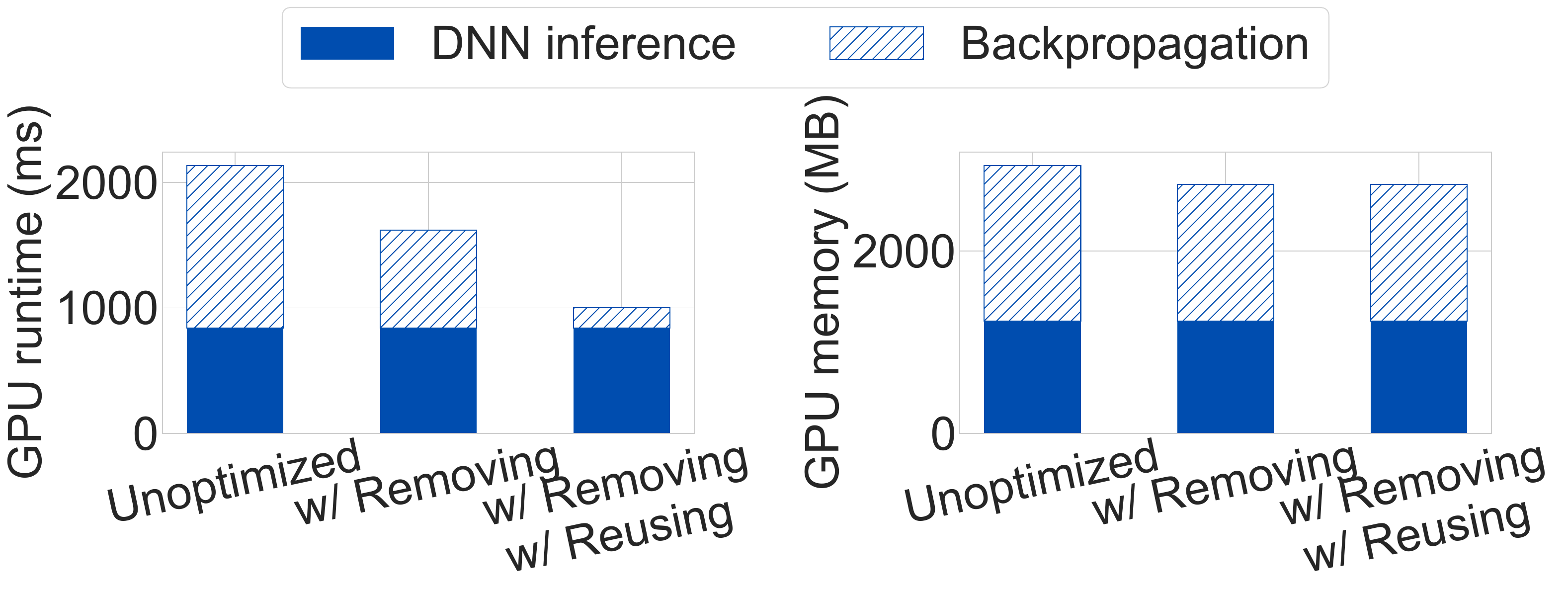}
    \vspace{-0.2cm}
    \tightcaption{Benchmarking the effectiveness of removing unneeded computation and DNNGrad reusing on a 10-frame video chunk using pipeline~\circled{a}. \name reduces the GPU runtime overhead by 87\% and the GPU memory overhead by 12\%.}
    \label{fig:GPU-overhead-reduction}
\end{figure}

\mypara{Summary}
After removing unneeded computation and \dnngrad reusing, \name's GPU computation overhead is below 20\% of DNN inference (as shown in Figure~\ref{fig:GPU-overhead-reduction}) and the GPU memory overhead is reduced by 12\%.

\tightsubsection{Reducing CPU overhead}
\label{subsec:cpu-overhead-reduction}


\name requires encoding the video multiple times to obtain \inputgrad for each knob, incurring high CPU overhead.

\name reduces such overhead to one single encoding for those knobs that operate on \textit{non-overlapping} spatial areas of DNN input 
\review{(note that \name only uses this technique on a specific set of knobs like the encoding qualities of different spatial blocks and it may not generalize to other knobs.).}{This addresses Reviewer B, C's concern on the generalizability of CPU optimization.}
For example, say the analytic system contains two knobs $k_1,k_2$, where $k_1$ is the encoding quality of the left half of the video, and $k_2$ is the encoding quality of the right half of the video.
As a result, $k_1$ mainly changes the left half of the input video, and $k_2$ mainly changes the right half of the video\footnote{$k_1$ may slightly change the right half of the video. However, such change is minor compared to how much $k_2$ changes the right half of the video}.
To calculate the \inputgrad of $k_1$ and $k_2$, We can change $k_1$ and $k_2$ \textit{simultaneously} and then use the change of the left half of the input video to compute the \inputgrad of $k_1$, and the right half to compute the \inputgrad of $k_2$.

\tightsubsection{Reducing network overhead}

\name imposes high network overhead as it streams the \dnngrad from the server back to the sensor.
This overhead can be substantial.
For instance, the \dnngrad of an object detector on a frame will be as large as the original raw, uncompressed frame size.




To compress \dnngrad, our observation is: modern codecs (\eg video codecs~\cite{h264,h265}) typically partition the data into small groups, called \textit{Minimum Coded Units} (MCUs), and then decide the compression scheme \textit{on top of} these groups.
In other words, the data inside each MCU will share the same compression scheme, thus there is no need to further distinguish between the data inside each MCU.
Based on this observation, for each MCU, \name takes the absolute value of \dnngrad and averages the values of each data inside the MCU.
For example, when the input video stream is encoded by H.264~\cite{h264} video codec, each MCU will be a $16\times 16$ pixel block.
\name then averages the \dnngrad values inside each $16\times16$ pixel block and obtains only one \dnngrad number for each pixel block, which compresses \dnngrad by $16^2\times$ and makes the bandwidth overhead of streaming \dnngrad negligible.
\review{
We note that this optimization does not apply to all existing codecs, but it does cover a wide range of existing codecs (\eg video codecs~\cite{h264,h265}, image codecs~\cite{jpeg} and audio codecs~\cite{audio-codec}) and it covers all codecs that \name is using in its evaluation.
}{This addresses Reviewer B, C's concern on the generality of network bandwidth optimization technique.}

\begin{table*}[]
\scriptsize
\setlength\tabcolsep{0.2em}
\centering

\begin{tabular}{|c|c|c|c|c|c|c|c|c|c|c|}
\hline
ID          & \begin{tabular}[c]{@{}c@{}}Target\\ application\end{tabular}                  & \begin{tabular}[c]{@{}c@{}}Analytic\\ task\end{tabular}                      & Dataset                                                   & \begin{tabular}[c]{@{}c@{}}Streaming\\media type\end{tabular}           & DNN                                                                     & \begin{tabular}[c]{@{}c@{}}Accuracy\\ metric\end{tabular} & Configuration knobs                                                                                                    & \begin{tabular}[c]{@{}c@{}}Trade-off \\ between \\ accuracy and\end{tabular} & Baselines                                                                                                           & \begin{tabular}[c]{@{}c@{}}Showing the \\ generalizability of\\\name across\end{tabular}            \\ \hline\hline
\circled{a} & \multirow{3}{*}{\begin{tabular}[c]{@{}c@{}}Autonomous\\ driving\end{tabular}} & \multirow{3}{*}{\begin{tabular}[c]{@{}c@{}}Vehicle\\ detection\end{tabular}} & \multirow{2}{*}{\sf{Downtown}~\cite{googleOneAdaptDriving}}                            & \multirow{3}{*}{RGB} & \multirow{3}{*}{\begin{tabular}[c]{@{}c@{}}Faster-\\ RCNN\end{tabular}} & \multirow{3}{*}{F1 score}                                 & \begin{tabular}[c]{@{}c@{}}Fine-grained encoding \\ quality assignment\end{tabular}                                    & \multirow{2}{*}{\begin{tabular}[c]{@{}c@{}}Bandwidth\\ usage\end{tabular}}   & \begin{tabular}[c]{@{}c@{}}DDS~\cite{dds} \\ EAAR~\cite{eaar}\\ AccMPEG~\cite{accmpeg}\end{tabular}                 & \multirow{3}{*}{\begin{tabular}[c]{@{}c@{}}Different\\ \textbf{configuration}\\\textbf{knobs}\end{tabular}}      \\ \cline{1-1} \cline{8-8} \cline{10-10}
\circled{b} &                                                                               &                                                                              &                                                           &                      &                                                                         &                                                           & Video codec knobs                                                                                                      &                                                                              & \begin{tabular}[c]{@{}c@{}}Chameleon~\cite{chameleon} \\ CASVA~\cite{casva}\\ AWStream~\cite{awstream}\end{tabular} &                                                                                                        \\ \cline{1-1} \cline{4-4} \cline{8-10}
\circled{c} &                                                                               &                                                                              & \sf{Country}~\cite{googleOneAdaptDriving}                                              &                      &                                                                         &                                                           & Frame filtering knobs                                                                                                  & \begin{tabular}[c]{@{}c@{}}GPU\\ computation\end{tabular}                    & \begin{tabular}[c]{@{}c@{}}Profiling~\cite{awstream,chameleon}\\ Reducto~\cite{reducto}\end{tabular}                &                                                                                                        \\ \hline
\circled{d} & \begin{tabular}[c]{@{}c@{}}Autonomous\\ driving\end{tabular}                  & \begin{tabular}[c]{@{}c@{}}Vehicle\\ detection\end{tabular}                  & KITTI~\cite{kitti}                                                     & LiDAR                & \begin{tabular}[c]{@{}c@{}}Point-\\ Pillars\end{tabular}                & F1 score                                                  & \begin{tabular}[c]{@{}c@{}}Fine-grained encoding \\ quality assignment\end{tabular}                                    & \begin{tabular}[c]{@{}c@{}}Bandwidth\\ usage\end{tabular}                    & \begin{tabular}[c]{@{}c@{}}Region-based~\cite{dds,eaar}\\ Uniform quality\end{tabular}                              & \multirow{4}{*}{\begin{tabular}[c]{@{}c@{}}Different types of\\ \textbf{streaming media}\end{tabular}} \\ \cline{1-10}
\circled{e} & \multirow{3}{*}{\begin{tabular}[c]{@{}c@{}}Home\\ security\end{tabular}}      & \multirow{3}{*}{\begin{tabular}[c]{@{}c@{}}Human\\ detection\end{tabular}}   & \multirow{3}{*}{PKU-MMD~\cite{pku-mmd}}                                  & \multirow{3}{*}{\begin{tabular}[c]{@{}c@{}}RGB,\\Depth,\\InfraRed\end{tabular}} & \multirow{3}{*}{Yolo}                                                   & \multirow{3}{*}{mIoU}                                     & \multirow{3}{*}{\begin{tabular}[c]{@{}c@{}}Fine-grained encoding \\ quality assignment on\\ DNN features\end{tabular}} & \multirow{3}{*}{\begin{tabular}[c]{@{}c@{}}Bandwidth\\ usage\end{tabular}}   & \multirow{3}{*}{Region-based~\cite{dds,eaar}}                                                                       &                                                                                                        \\ \cline{1-1}
\circled{f} &                                                                               &                                                                              &                                                           &                      &                                                                         &                                                           &                                                                                                                        &                                                                              &                                                                                                                     &                                                                                                        \\ \cline{1-1}
\circled{g} &                                                                               &                                                                              &                                                           &                      &                                                                         &                                                           &                                                                                                                        &                                                                              &                                                                                                                     &                                                                                                        \\ \hline
\circled{h} & \begin{tabular}[c]{@{}c@{}}Smart\\ home\end{tabular}                          & \begin{tabular}[c]{@{}c@{}}Audio-\\ to-text\end{tabular}                     & \begin{tabular}[c]{@{}c@{}}Google\\ AudioSet~\cite{audioset}\end{tabular} & Audio                & Wav2Vec                                                                 & F1 score                                                  & Audio sampling rate                                                                                                    & \begin{tabular}[c]{@{}c@{}}Bandwidth\\ usage\end{tabular}                    & \begin{tabular}[c]{@{}c@{}}Profiling~\cite{awstream,chameleon}\\ Voice detection~\cite{stft}\end{tabular}           & \multirow{2}{*}{\begin{tabular}[c]{@{}c@{}}Different\\ \textbf{analytic tasks}\end{tabular}}           \\ \cline{1-10}
\circled{i} & \begin{tabular}[c]{@{}c@{}}Traffic\\ analytics\end{tabular}                   & \begin{tabular}[c]{@{}c@{}}Vehicle\\ segmentation\end{tabular}               & \sf{Traffic}~\cite{googleOneAdaptDriving}                                              & RGB                  & \begin{tabular}[c]{@{}c@{}}Mask-\\ RCNN\end{tabular}                    & F1 score                                                  & \begin{tabular}[c]{@{}c@{}}Fine-grained encoding \\ quality assignment\end{tabular}                                    & \begin{tabular}[c]{@{}c@{}}Bandwidth\\ usage\end{tabular}                    & \begin{tabular}[c]{@{}c@{}}DDS~\cite{dds}\\ AccMPEG~\cite{accmpeg}\end{tabular}                                     &                                                                                                        \\ \hline
\end{tabular}

\vspace{0.2cm}
    \tightcaption{\review{Overview of the experimental setup. Each row represents an analytic pipeline we used to evaluate \name.}{This addresses Reviewer B, C, and D's comments on experimental settings.}}
    \label{tab:pipeline}
\end{table*}

\tightsection{Evaluation}
\label{sec:eval}

\begin{figure}[t!]
    \centering

    \vspace{-0.1cm}
    \subfloat[][Pipeline \circled{a}]
    {
        \includegraphics[width=0.245\textwidth]{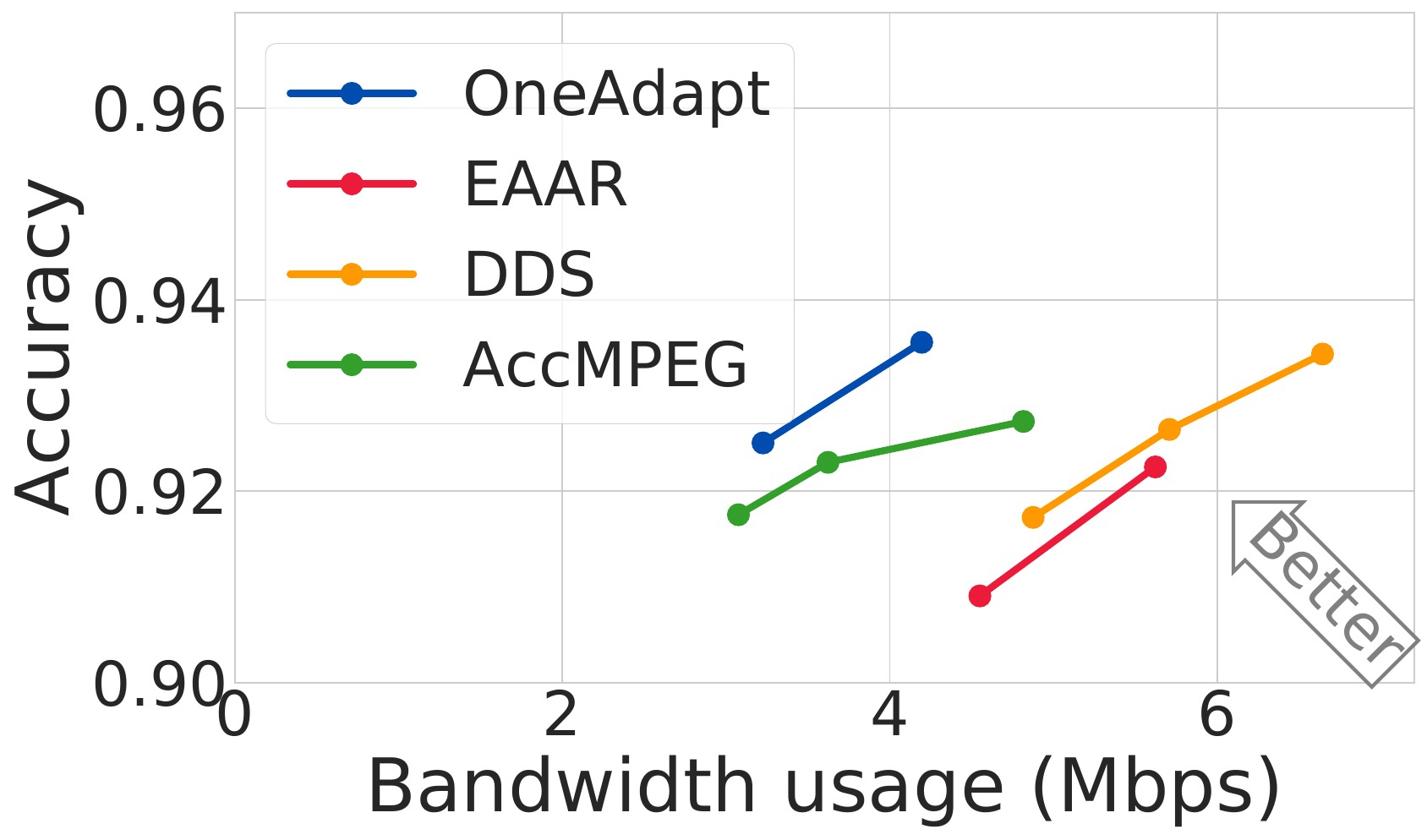} 
    }
    \subfloat[][Pipeline \circled{b}] 
    {
        \includegraphics[width=0.245\textwidth]{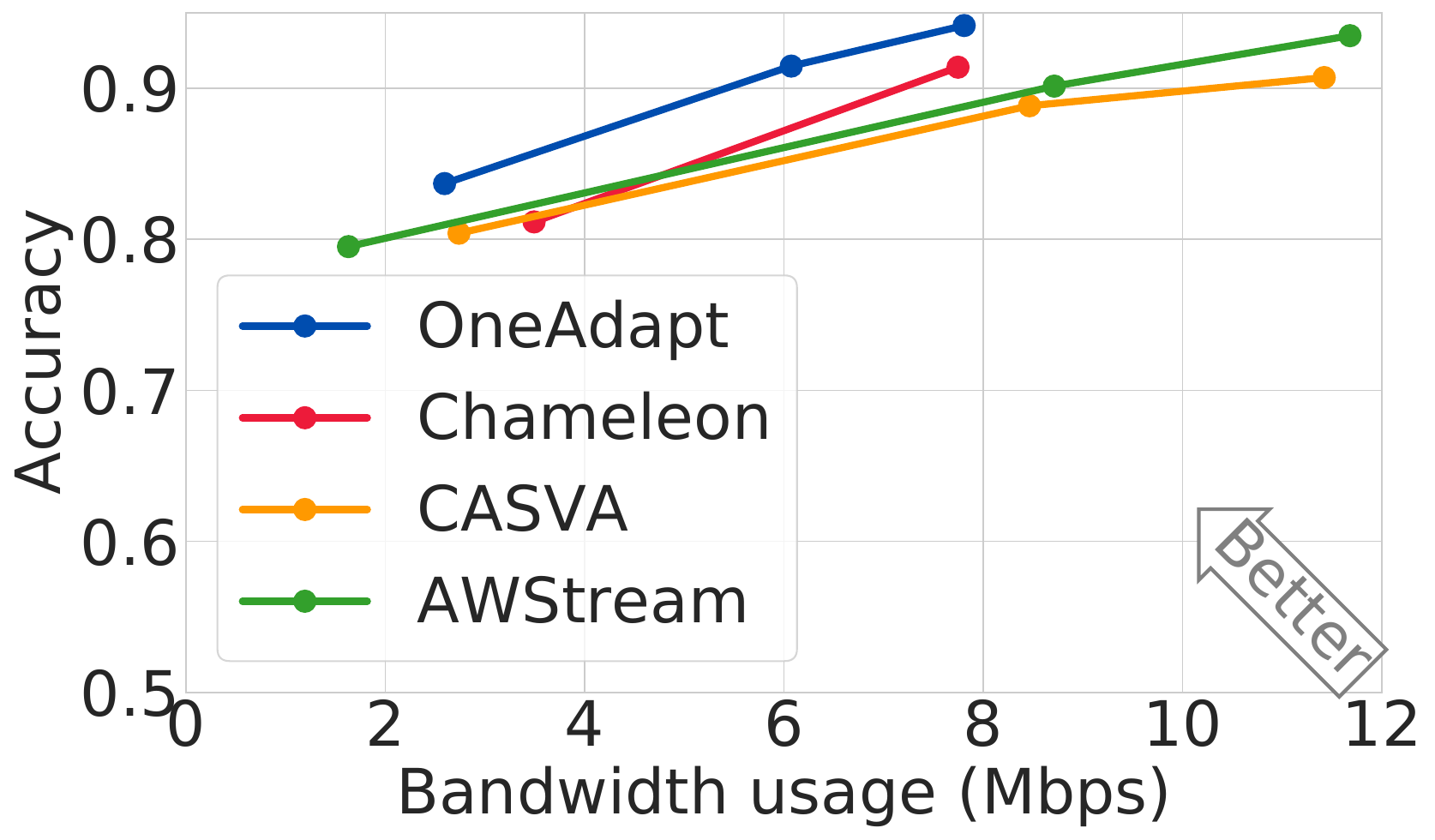} 
    }

    \vspace{-0.1cm}
    \subfloat[][Pipeline \circled{c}] 
    {
        \includegraphics[width=0.245\textwidth]{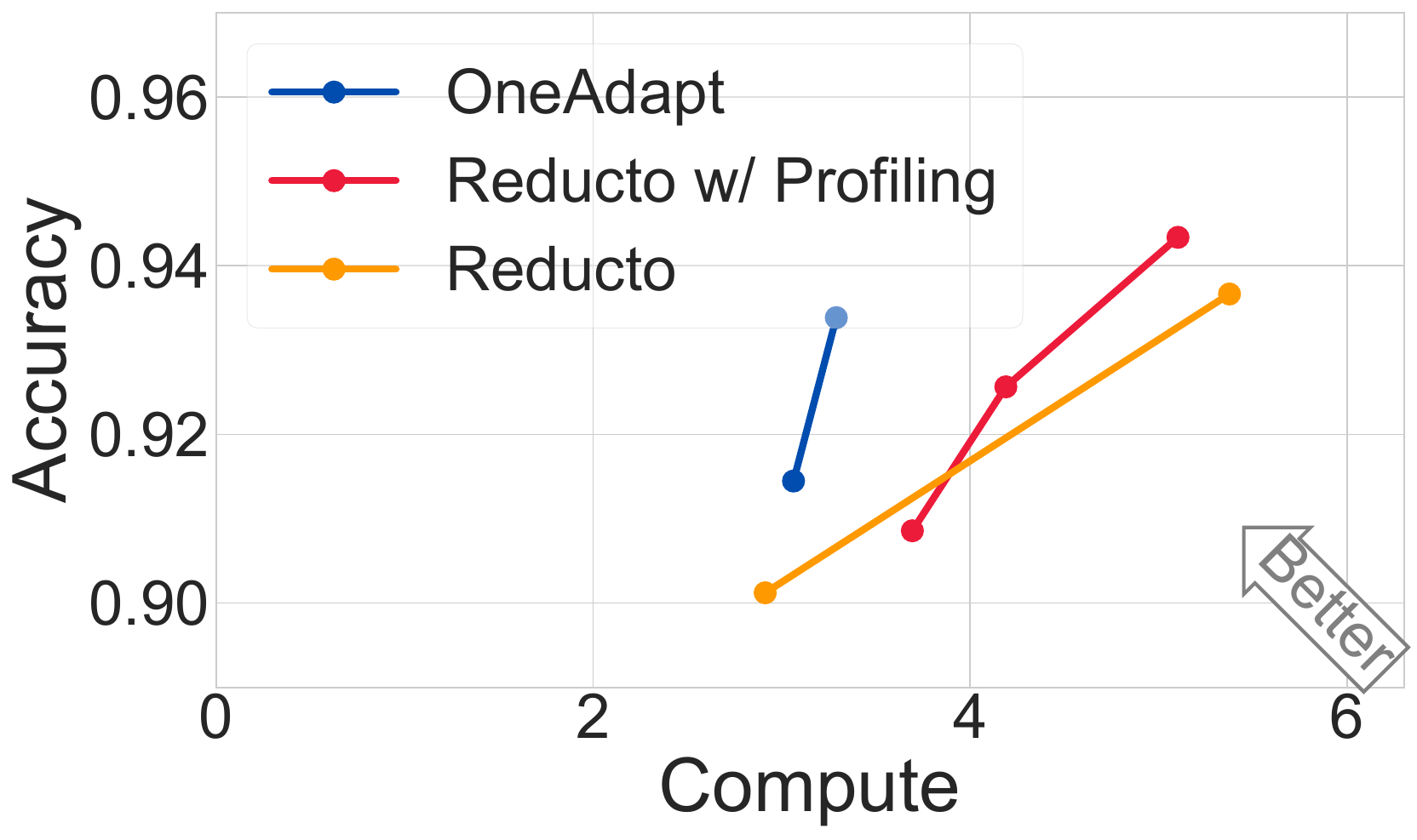} 
    }
    \subfloat[][Pipeline \circled{d}]  
    {
        \includegraphics[width=0.245\textwidth]{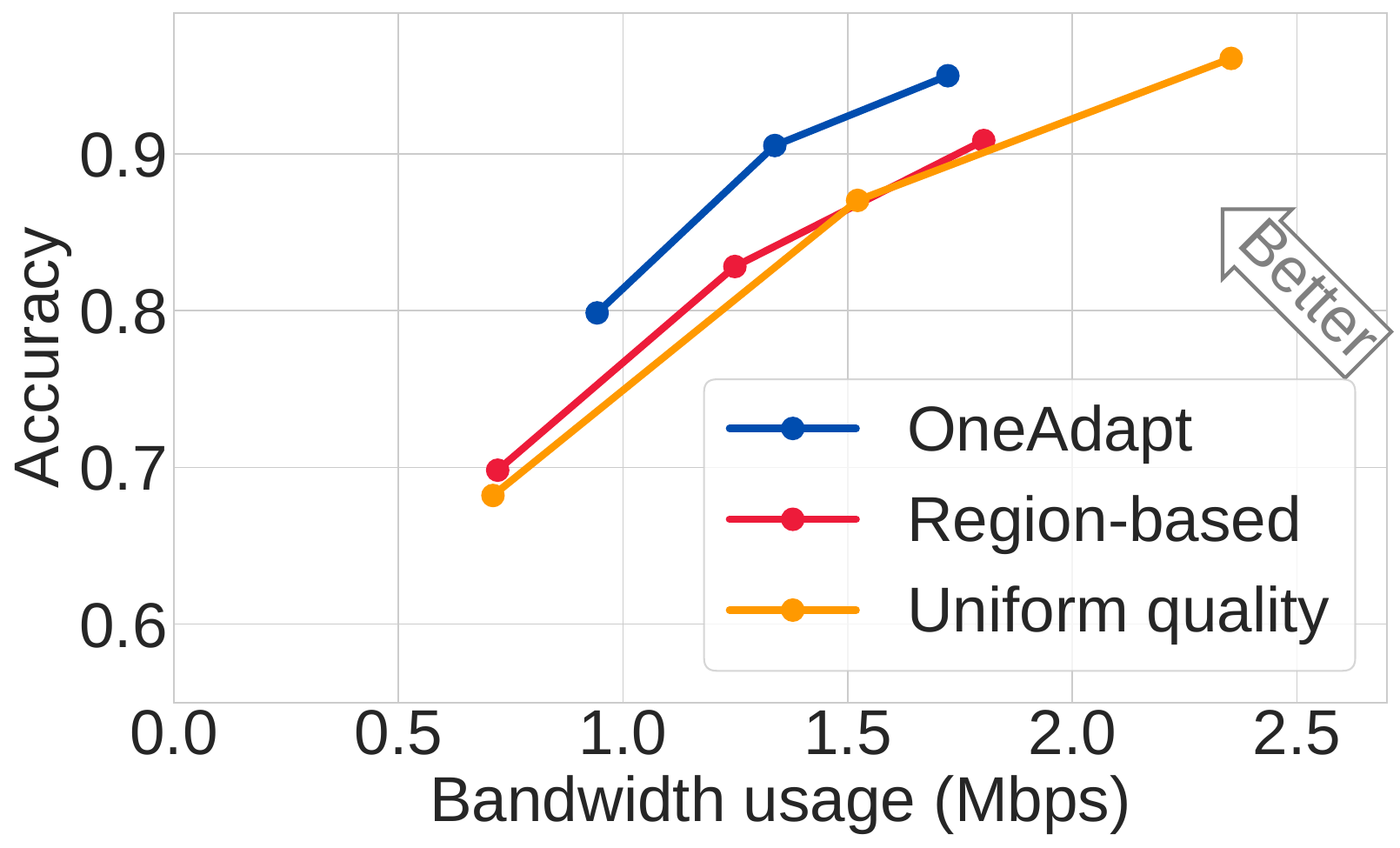} 
    }

    \vspace{-0.1cm}
    \subfloat[][Pipeline \circled{e}]  
    {
        \includegraphics[width=0.245\textwidth]{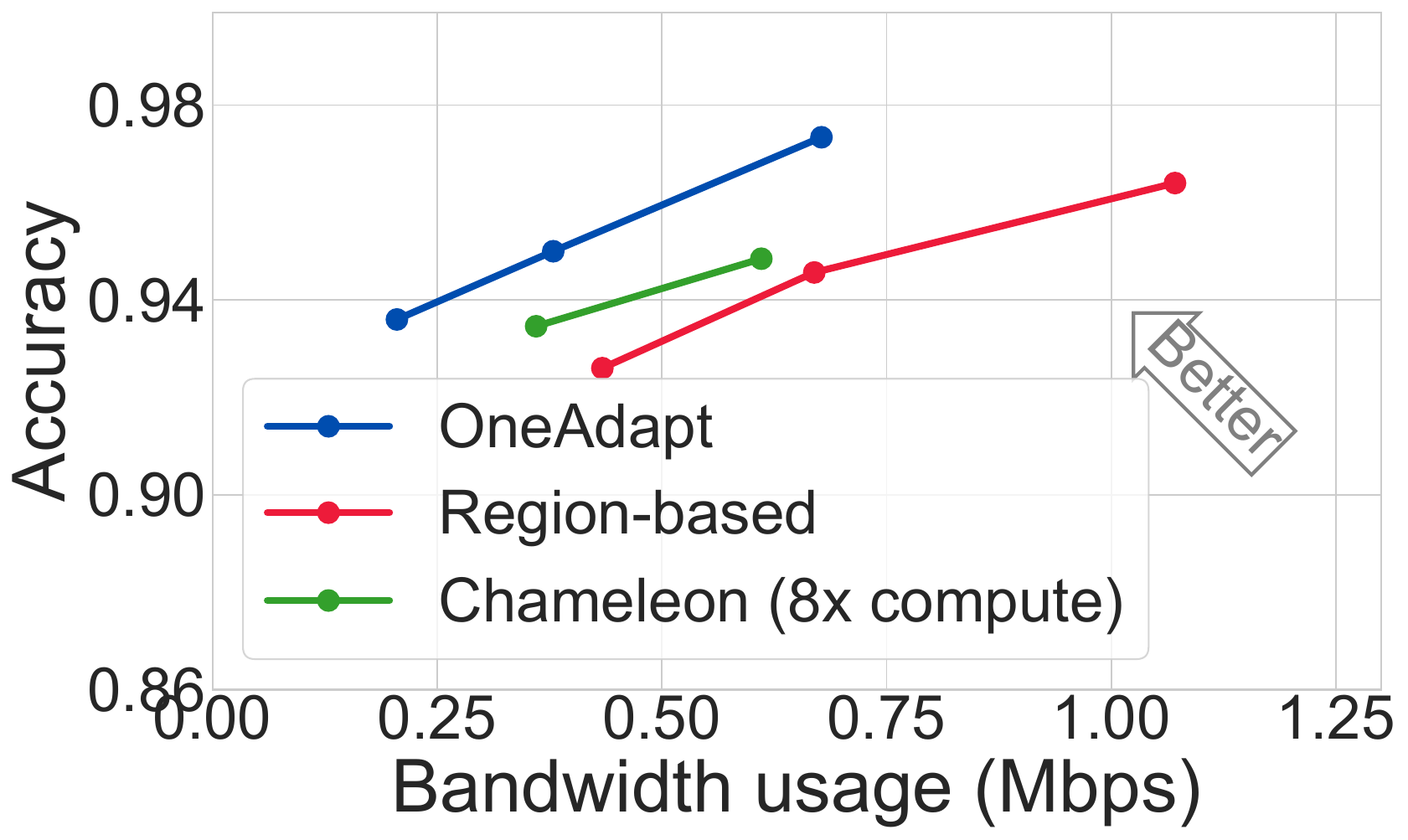} 
        \label{subfig:compute-improvement}
    }
    \subfloat[][Pipeline \circled{f}]  
    {
        \includegraphics[width=0.245\textwidth]{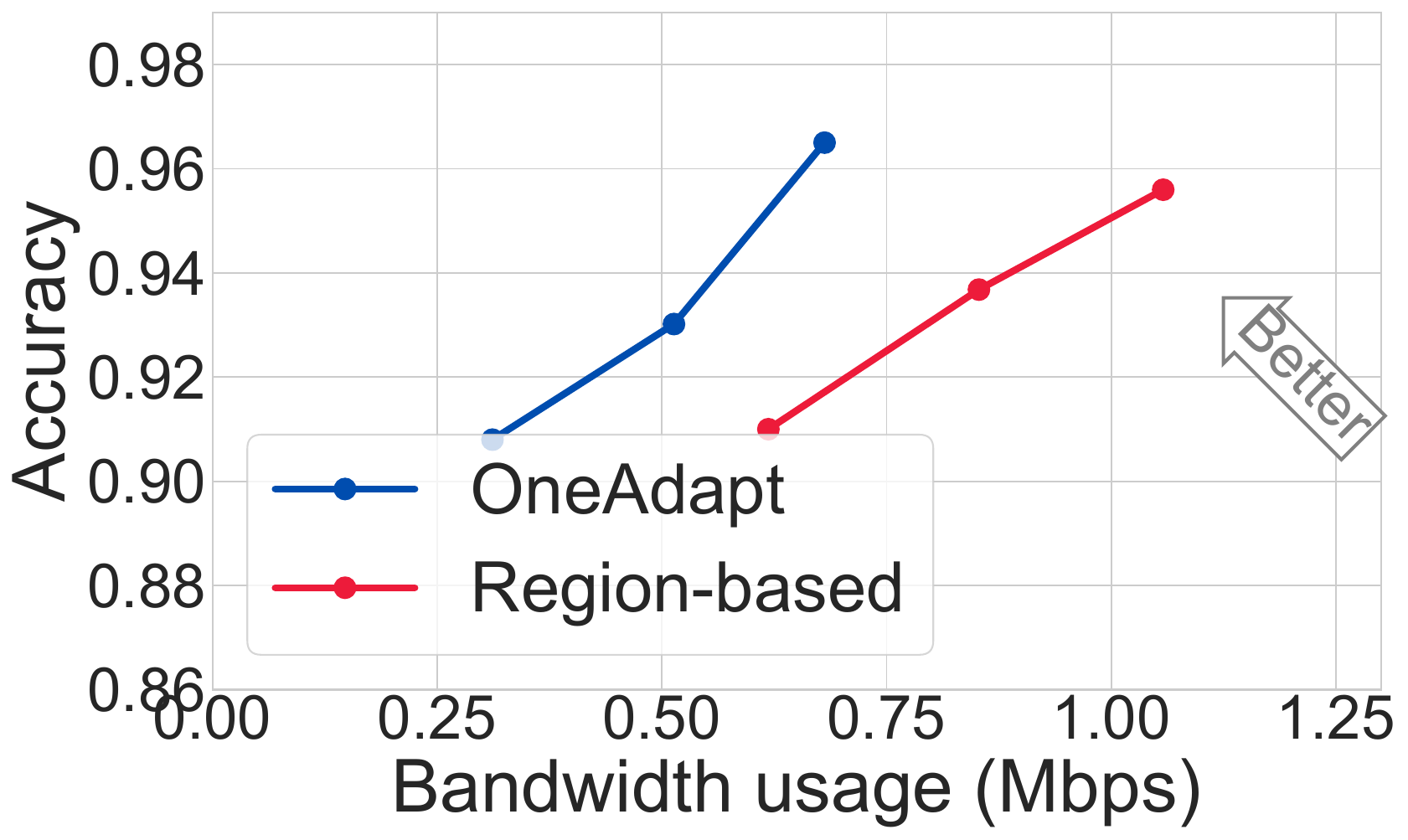}
    }
    
    \vspace{-0.1cm}
    \subfloat[][Pipeline \circled{g}]  
    {
        \includegraphics[width=0.245\textwidth]{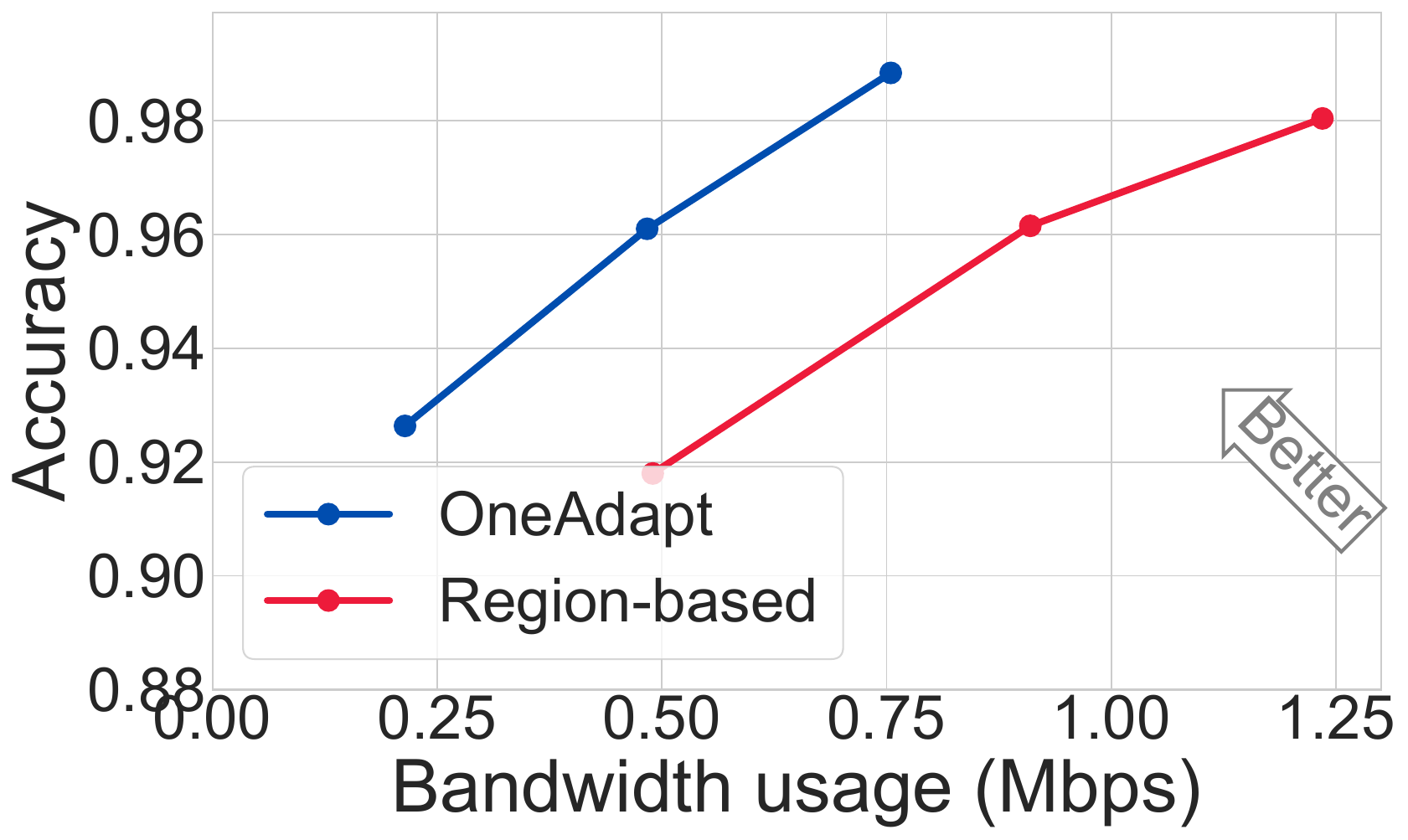} 
    }
     \subfloat[][Pipeline \circled{h}]  
    {
        \includegraphics[width=0.245\textwidth]{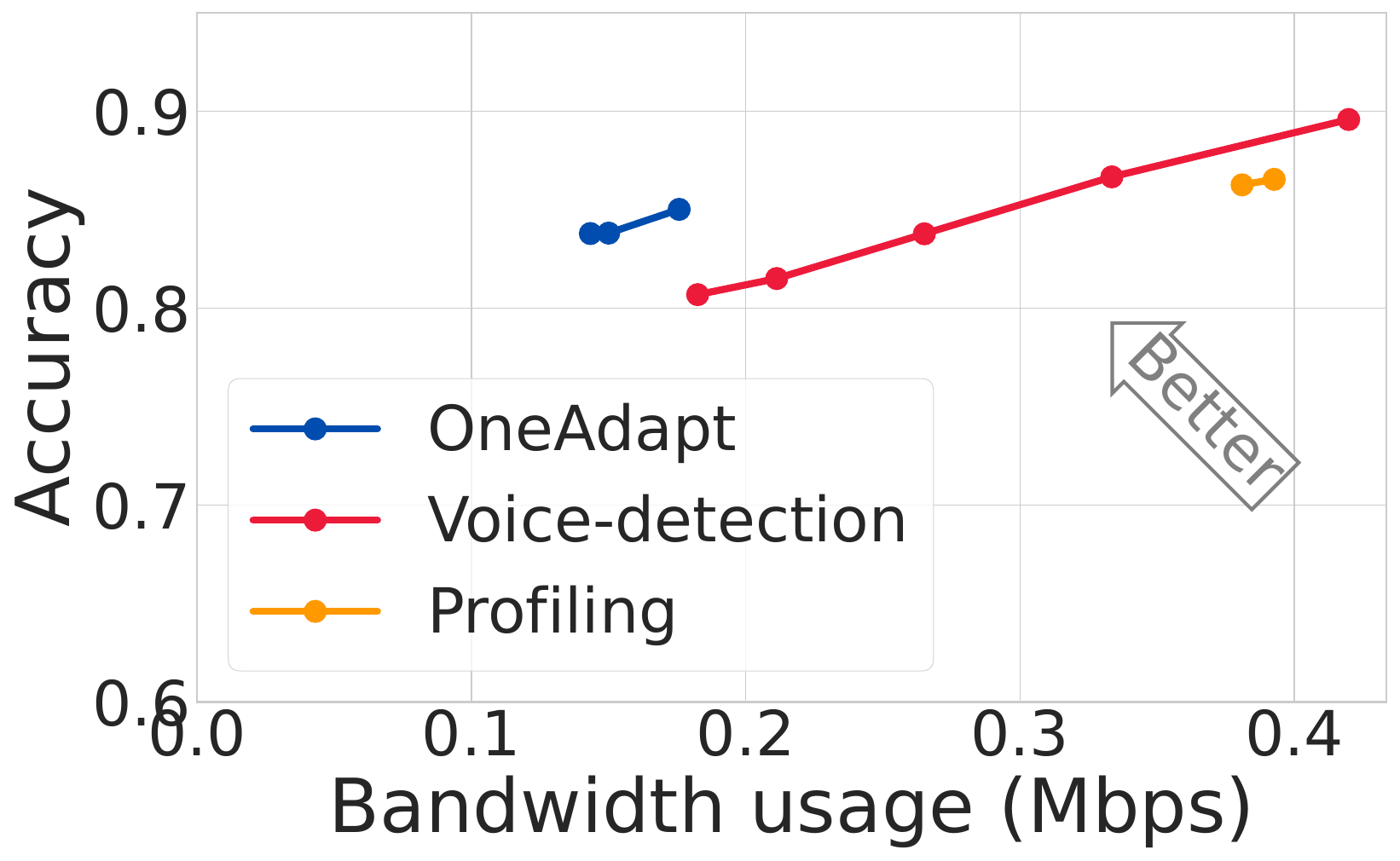} 
    }
    
    \vspace{-0.1cm}
    \subfloat[][Pipeline \circled{i}]  
    {
        \includegraphics[width=0.245\textwidth]{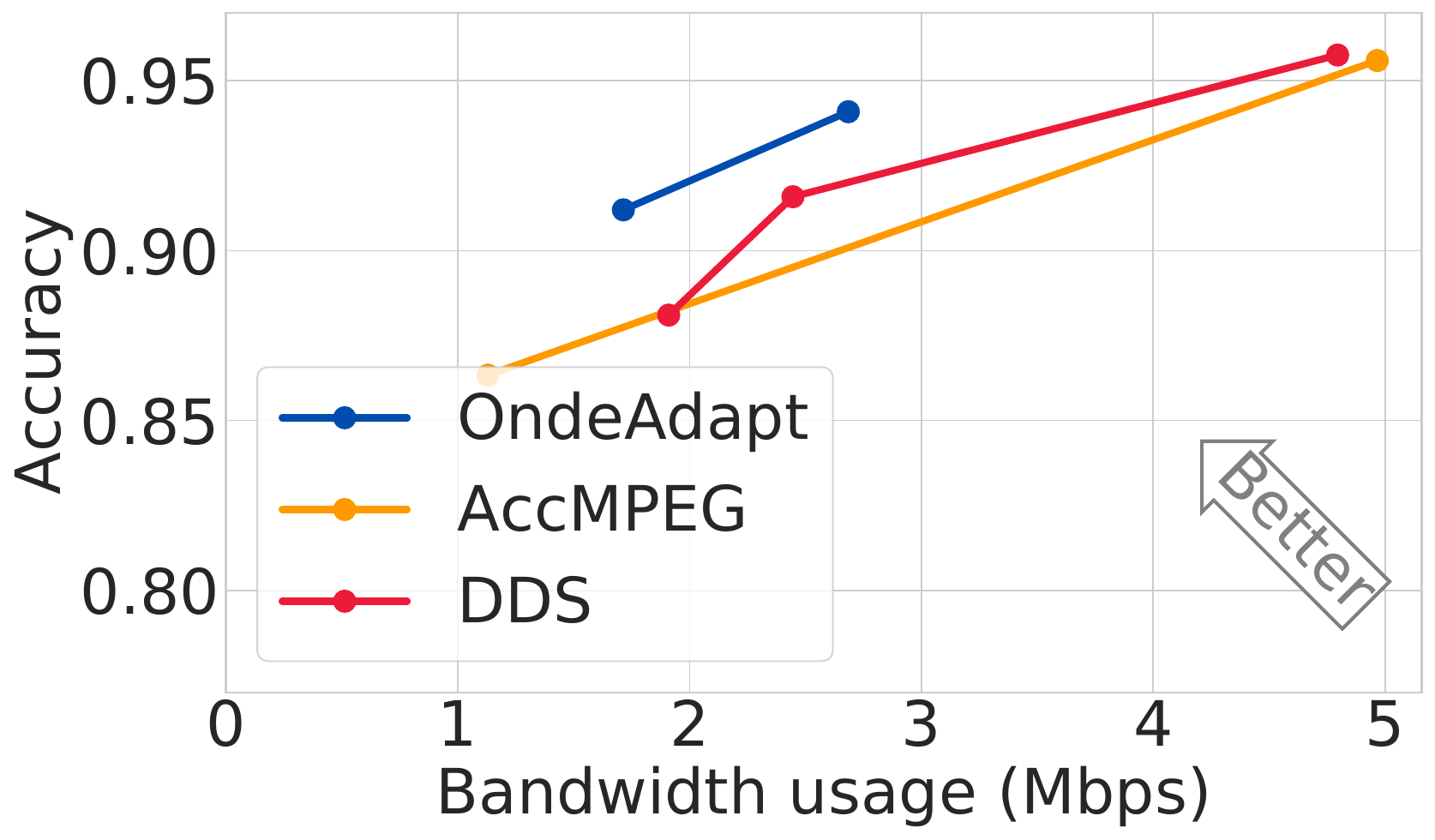} 
    }
    \vspace{-0.1cm}
    \tightcaption{
    Demonstrating the trade-off between accuracy and resource usage of of \name and several baselines. 
    \name achieves higher accuracy with 15-59\% resource usage reduction or 1-5\% higher accuracy with less resource usage compared to the baselines on 9 different pipelines.
    We note that the results in each figure are averaged across 5-10 videos or 200 audios (depending on the dataset used for each figure) and thus have statistical confidence.
    }
    \label{fig:eval-overall-resource-consumption}
\end{figure}

In our evaluation, we show that:
\begin{packeditemize}
\item \name achieves similar accuracy while reducing resource usage by 15-59\% or improves accuracy by 1-5\%  with the same or less resource usage when compared to state-of-the-art adaptation approaches. 
\review{This accuracy improvement is substantial, as the system needs to consume much more resources to improve accuracy by 1-5\% (\eg 2x more bandwidth (Figure~\ref{fig:eval-overall-resource-consumption}g) only improves 4\% accuracy). This improvement is also on par with prior work~\cite{dds,accmpeg}.}{This addresses Reviewer C's concern on whether the accuracy improvement is substantial}
\item The extra GPU overhead caused by \name is comparable to or lower than existing adaptation approaches.
\item \name achieves more resource reduction when there are more configuration knobs to tune.
\end{packeditemize}

Our code is available in~\cite{oneadapt-repo}.

\tightsubsection{Evaluation setup}
\label{subsec:eval-setup}

\mypara{Applications and  DNNs}
\review{We target four types of applications: vehicle detection with FasterRCNN for autonomous driving~\cite{faster-rcnn,detectron2}, vehicle segmentation employing MaskRCNN for traffic analytics~\cite{detectron2,maskrcnn}, human detection using YoLo for home security~\cite{yolo,yolo-pytorch} and audio-to-text utilizing Wav2Vec for smart home applications~\cite{wav2vec,wav2vec-github}.
Note that both detection and segmentation applications categorize objects as either of interest or not.
}{This addresses Reviewer C's concern on whether the applications perform binary- or multi-class classification}

\mypara{Accuracy metrics}
We use F1 score~\cite{dds,awstream,chameleon,casva} for vehicle detection, vehicle segmentation, and audio-to-text.
For human detection, we use mean IoU (mIoU~\cite{eaar}).
\review{Following prior work~\cite{awstream,chameleon,dds,accmpeg,ekya,boggart}, we define accuracy as the similarity between the current inference results and the inference results generated from the most resource-consuming configuration to measure the impact of adapting configurations on inference results.}{This addresses Reviewer C's concern on the ground truth of our evaluation}

\mypara{Input settings}
For all our applications we use 10FPS from the respective sensors (except for audio-to-text, where we chunk the raw audio into one-second segments and send them to the DNN for analytics). 
Note that while 30FPS and 60FPS are typical for video content intended for human viewing~\cite{video-fps}, 10FPS is prevalent in real-time analytic applications such as autonomous driving~\cite{autonomous-driving-frame-rate,autonomous-driving-frame-rate2,autonomous-driving-frame-rate3,autonomous-driving-frame-rate4}.

\mypara{Dataset}
We use the following datasets to evaluate \name, with the goal of covering various application scenarios and streaming media (summarized in Table~\ref{tab:dataset}):
\begin{packeditemize}
\item \textit{Autonomous driving}: our dataset covers two driving contexts (downtown driving and country driving) and two main types of sensors (RGB video sensor and LiDAR sensor).
Specifically, we obtain 10 downtown driving RGB videos~\cite{googleOneAdaptDriving} and 8 country driving RGB videos~\cite{googleOneAdaptDriving} using an anonymous YouTube search, and 6 urban driving LiDAR videos from KITTI dataset~\cite{kitti}. 
\item \textit{Traffic analytics:} We collect 5 traffic camera video footag by anonymous YouTube searche~\cite{googleOneAdaptDriving}.
\item \textit{Home security:} 
We use three types of sensor data (RGB video sensor, InfraRed sensor, and depth sensor), with 10 videos each from PKU-MMD dataset~\cite{pku-mmd}.
The inclusion of InfraRed and depth data is vital for enhancing night-time human detection accuracy in home security applications.
\item \textit{Smart home:} We randomly sample 200 audio clips in Google AudioSet~\cite{audioset}.
\end{packeditemize}




\review{
\mypara{Pipelines}
We use nine analytic pipelines (pipeline \circled{a}-\circled{i}).
These analytic pipelines show the applicability of  \name across configuration knobs (pipeline \circled{a}-\circled{c}), types of streaming media (pipeline \circled{d}-\circled{g}) and applications (pipeline \circled{h}-\circled{i}).
Table~\ref{tab:pipeline} summarizes these pipelines, including their target applications, analytic tasks, datasets, streaming media types, DNNs, and accuracy metrics.

\mypara{Knobs and baselines}
We describe the knobs and the corresponding baselines of these pipelines one by one:
\begin{packeditemize}
    \item Pipeline~\circled{a}: This pipeline saves bandwidth by adjusting the encoding quality within each 16x16 pixel macroblocks~\cite{h264,accmpeg}. We benchmark against three methods.
    DDS~\cite{dds} uses a low-quality video for initial inference, then refines specific regions with high-quality encoding.
    EAAR~\cite{eaar} uses results from the previous frame to identify current frame regions needing high-quality encoding.
    AccMPEG~\cite{accmpeg} runs a sensor-side neural network to determine regions for high-quality encoding.
    \item Pipeline~\circled{b}: This pipeline optimizes bandwidth using three knobs: resolution, QP, and B-frame selection likelihood~\cite{awstream,chameleon,vstore,videostorm,dds,eaar,accmpeg,fischer2021saliency,vigil,reducto,filterforward,glimpse,casva}), all supported in prevalent video codecs~\cite{ffmpeg-options,h264,h265}.
    We implement three baselines for this pipeline.
    Chameleon~\cite{chameleon} periodically profiles top-k configurations and picks the one with the highest accuracy under the current bandwidth budget.
    AWStream~\cite{awstream} periodically profiles a downsampled subset of configurations and chooses the one that maximizes accuracy within bandwidth limits.
    CASVA~\cite{casva} runs a reinforcement learning model on the sensor to determine the new configuration. 
    \item Pipeline~\circled{c}: This pipeline reduces the GPU computation usage by running frame filtering using two frame filtering knobs (pixel difference threshold and area difference threshold~\cite{reducto}).
    As for baselines, we implement Reducto~\cite{reducto}, together with a profiling-based baseline~\cite{awstream}.
    \item Pipeline~\circled{d}: This pipeline saves bandwidth when streaming LiDAR point clouds by segmenting the 3D space around the LiDAR sensor into spatial blocks and streaming k\% of LiDAR points from each. 
    The value of $k$ can vary between blocks.
    While this encoding mechanism is basic, our goal is not to establish a best practice, but to showcase the advantage of configuration adaptation.
    To ensure robustness, we average results from five repeated experiments.
    We compare against two baselines: a region-based method extended from~\cite{eaar} and a uniform-quality approach, which applies a fixed encoding quality across blocks.
    \item Pipeline~\circled{e}\circled{f}\circled{g}: 
    These pipelines transform videos into feature vectors using a sensor-side DNN and then stream them for server analysis~\cite{vcm}. To reduce bandwidth usage, we vary encoding qualities across the spatial blocks of feature vectors.
    Traditional heuristics are not directly applicable on DNN-generated features, so we introduce a region-based heuristic that prioritizes blocks with recent human activity~\cite{dds,eaar}.
    Note that given the 16 configuration knobs in this pipeline, profiling methods fall short, even when configurations are aggressively downsampled. As evidence, we implement Chameleon~\cite{chameleon} in pipeline~\circled{e} as the profiling baseline, which, despite 8x more GPU resources than \name and only tuning 4 knobs, still fails to outperform \name.
    \item Pipeline~\circled{h}: This pipeline optimizes the delay between a user's speech and its transcription into text by reducing bandwidth usage.
    We use the audio sampling rate as our knob. We implement a profiling-based baseline (Chameleon~\cite{chameleon}) and a heuristics baseline that raises the audio sampling rate when detecting human voice~\cite{stft}.
    \item Pipeline~\circled{i}: This pipeline aims to reduce the bandwidth usage of running traffic analytics, by assigning different encoding quality to different spatial areas.
    We use the baselines same as pipeline~\circled{a} except for EAAR, as EAAR is not directly applicable to vehicle segmentation DNNs.

\end{packeditemize}
}{This addresses Reviewer B's comments on clarifying eval setup}

\mypara{Resource usage}
For those pipelines that aim to minimize bandwidth usage, we measure the bandwidth usage by the bandwidth needed to send one-second worth of data, while constraining the GPU computation as being able to analyze 1.5-second worth of data per second (we relax this constraint for one baseline (DDS~\cite{dds}) as it needs to examine the same one-second data twice for each second).
For those pipelines that aim to minimize GPU computation usage, we measure the GPU computation by the number of video frames that need to be analyzed per second\footnote{The backpropagation overhead of \name is also transformed to the number of frames analyzed per second (by using backpropagation runtime divided by the runtime of analyzing one frame).}) and we do not constraint the bandwidth usage of \name and other baselines.

\mypara{Hardware settings}
We use one Intel Xeon 4100 Silver CPU as the CPU and NVIDIA RTX 2080 as the GPU.




\tightsubsection{Experimental results}

\begin{figure}
    \centering
    
    
        \includegraphics[width=0.7\columnwidth]{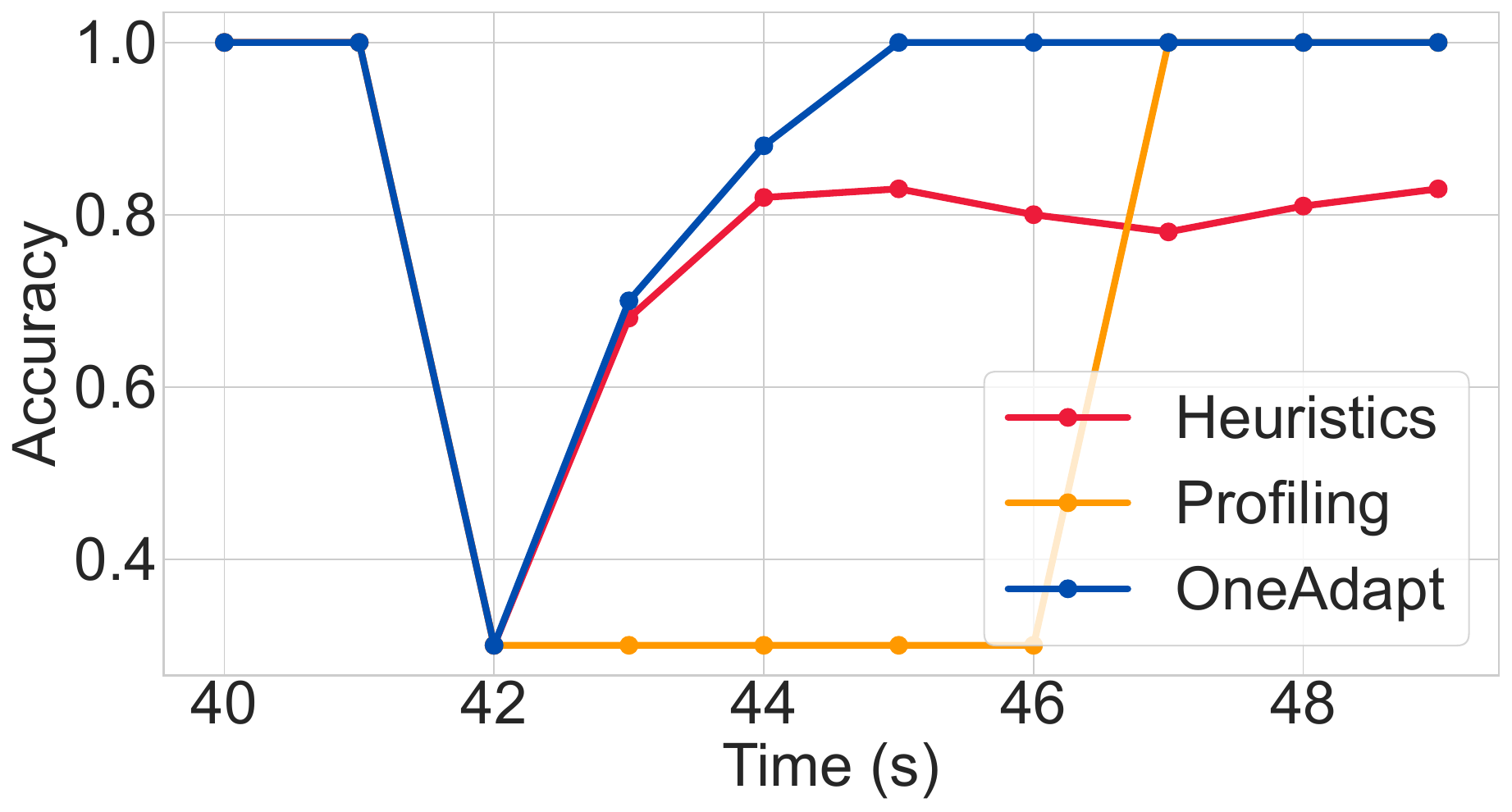}
     \tightcaption{Comparing the behavior of \name against the profiling-based baseline and heuristic-based baseline. The accuracy of \name and all baselines drop upon the person starts talking, but \name quickly adapts to the audio content change and improves its accuracy, while the accuracy of the baselines stays low for a while, as it profiles every 5 seconds.}

    \label{fig:adaptivity}
\end{figure}


\begin{figure}[t]
    \centering
    \includegraphics[width=0.7\columnwidth]{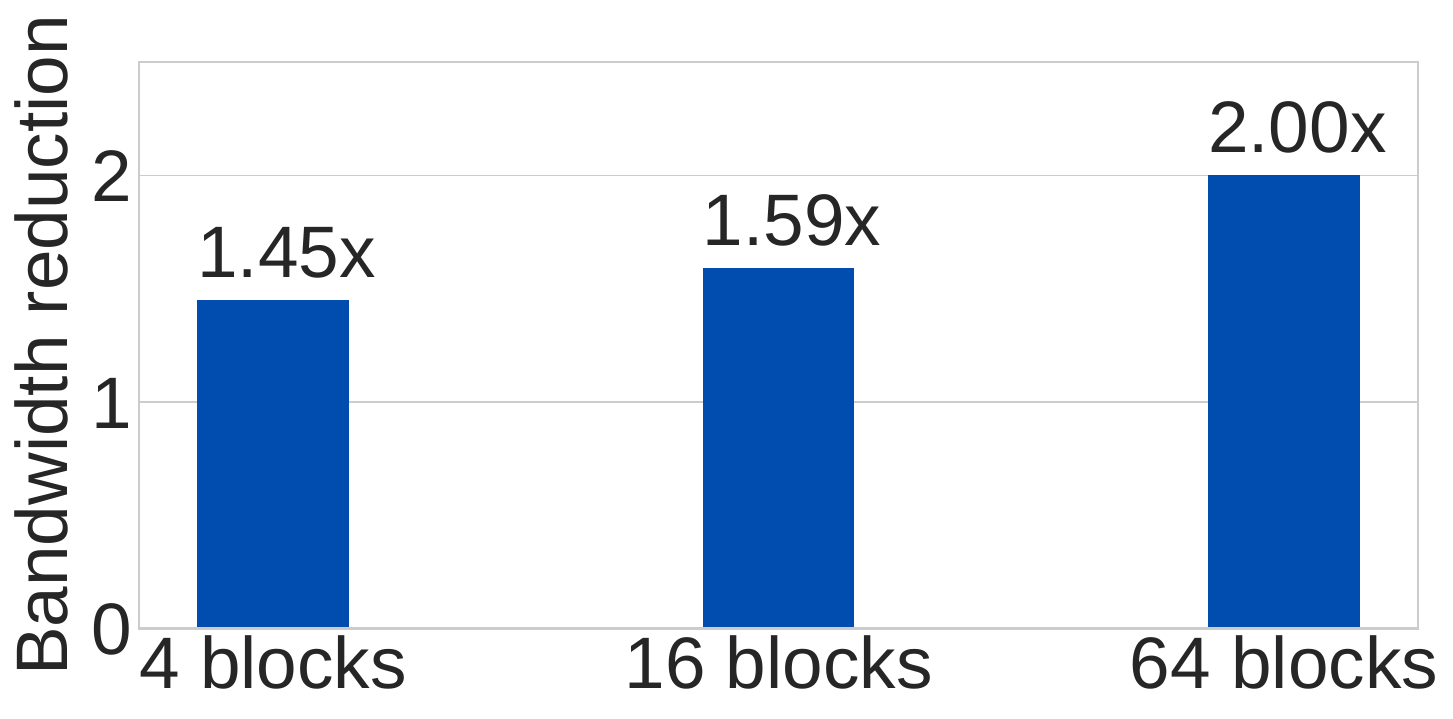}
    
    \tightcaption{When the knobs are more fine-grained, the bandwidth reduction of \name (the bandwidth usage of the region-based baseline, divided by the bandwidth usage of \name when \name is of higher accuracy) grows larger.}
    \label{fig:fine-grained-knobs}
\end{figure}

\begin{figure}[t!]
    \centering
    \subfloat[][Pipeline \circled{a}\circled{i}  GPU compute]
    {
        \includegraphics[width=0.49\columnwidth]{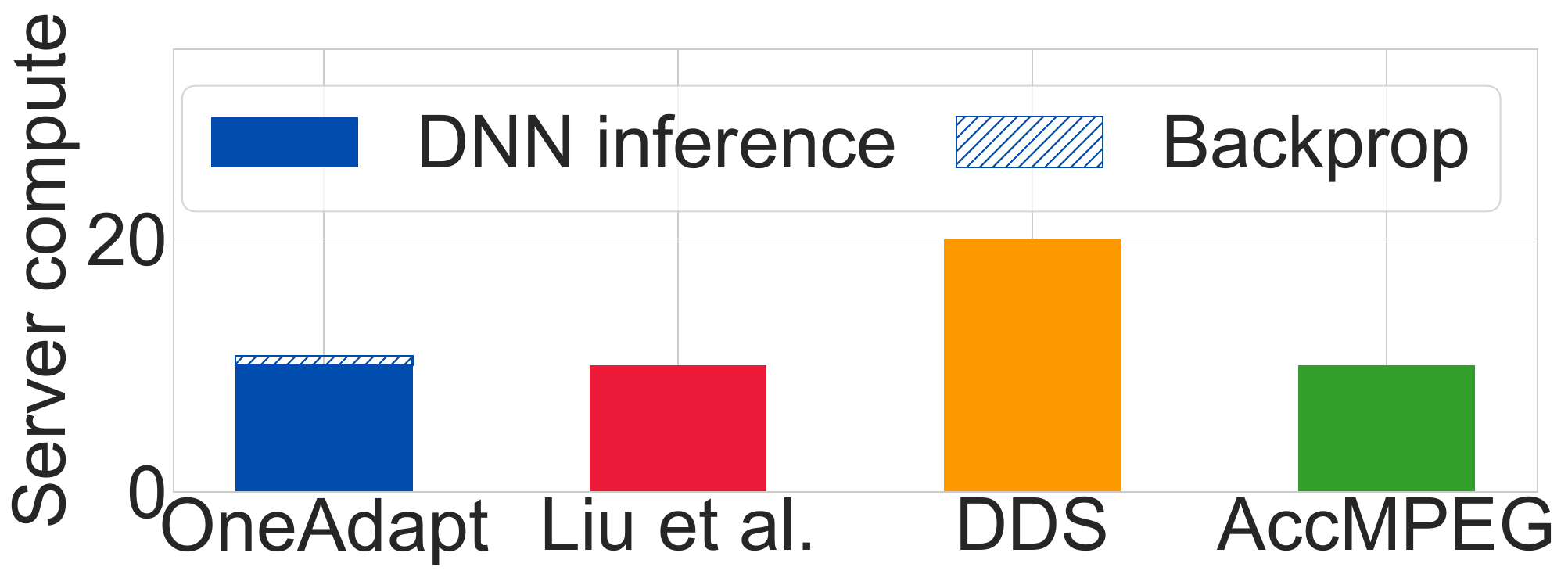} 
        \label{subfig:serv1}
    }
    \subfloat[][Pipeline \circled{b} GPU compute]
    {
        \includegraphics[width=0.49\columnwidth]{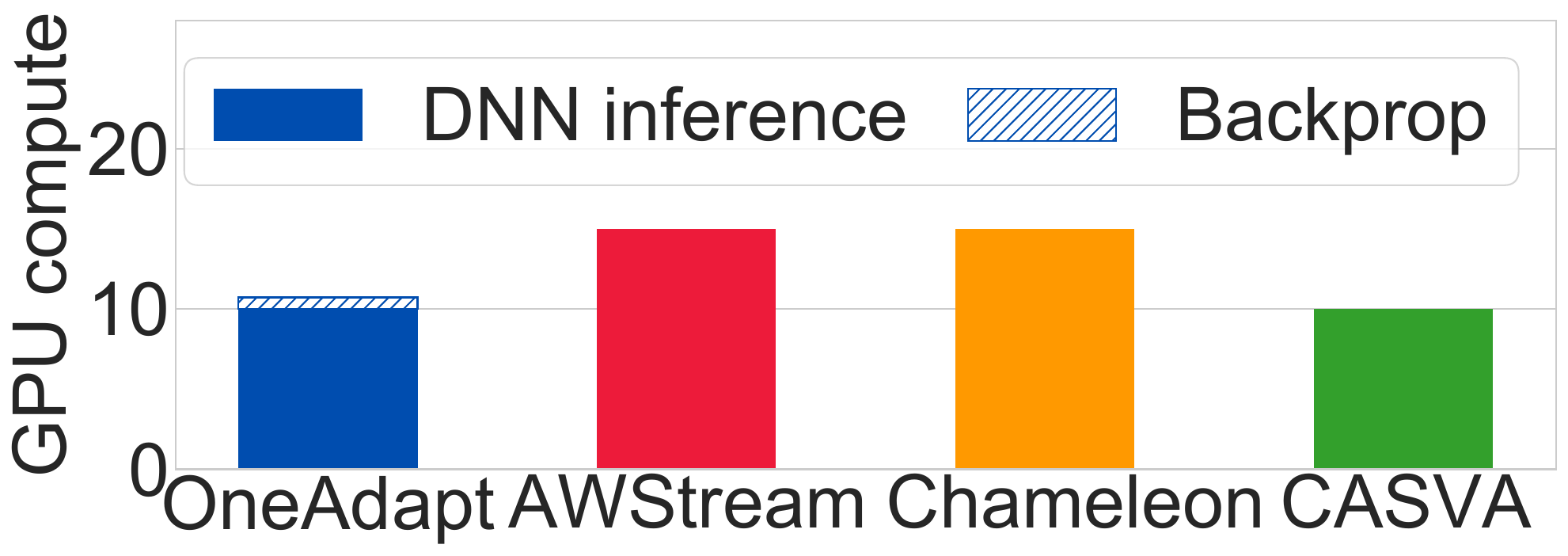} 
        \label{subfig:serv2}
    }
    
    \subfloat[][Pipeline \circled{c}  GPU compute]
    {
        \includegraphics[width=0.49\columnwidth]{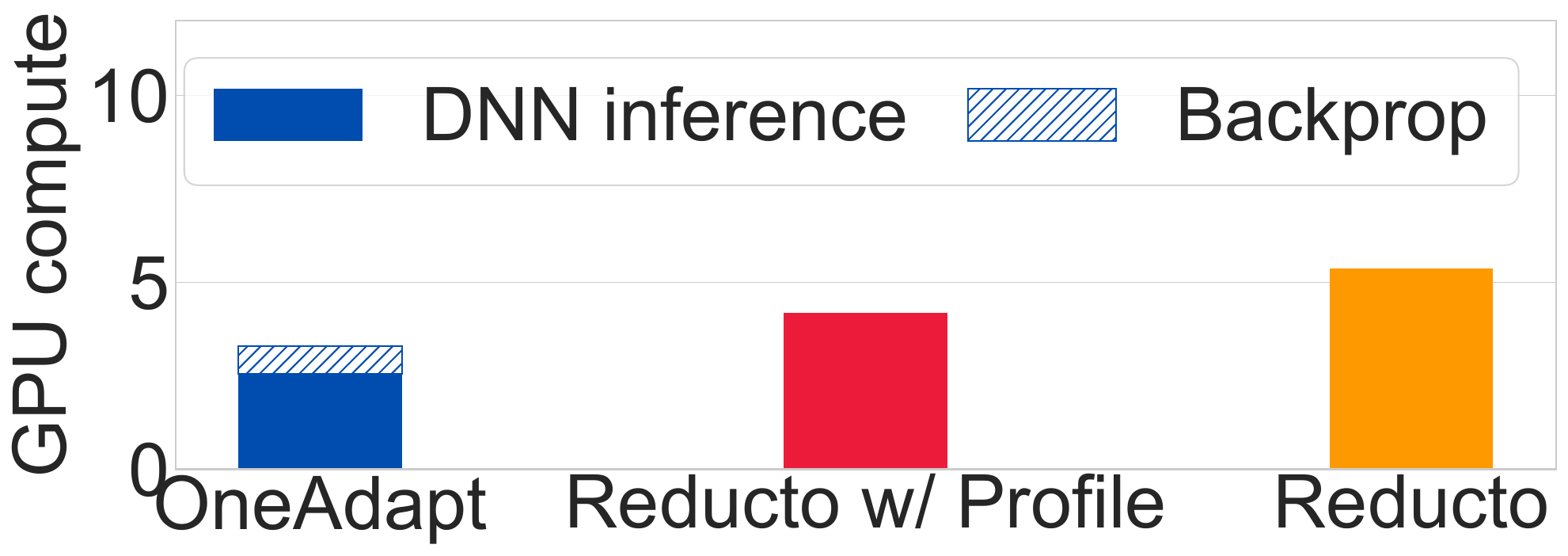} 
        \label{subfig:serv3}
    }
    \subfloat[][Pipeline \circled{d}  GPU compute]
    {
        \includegraphics[width=0.49\columnwidth]{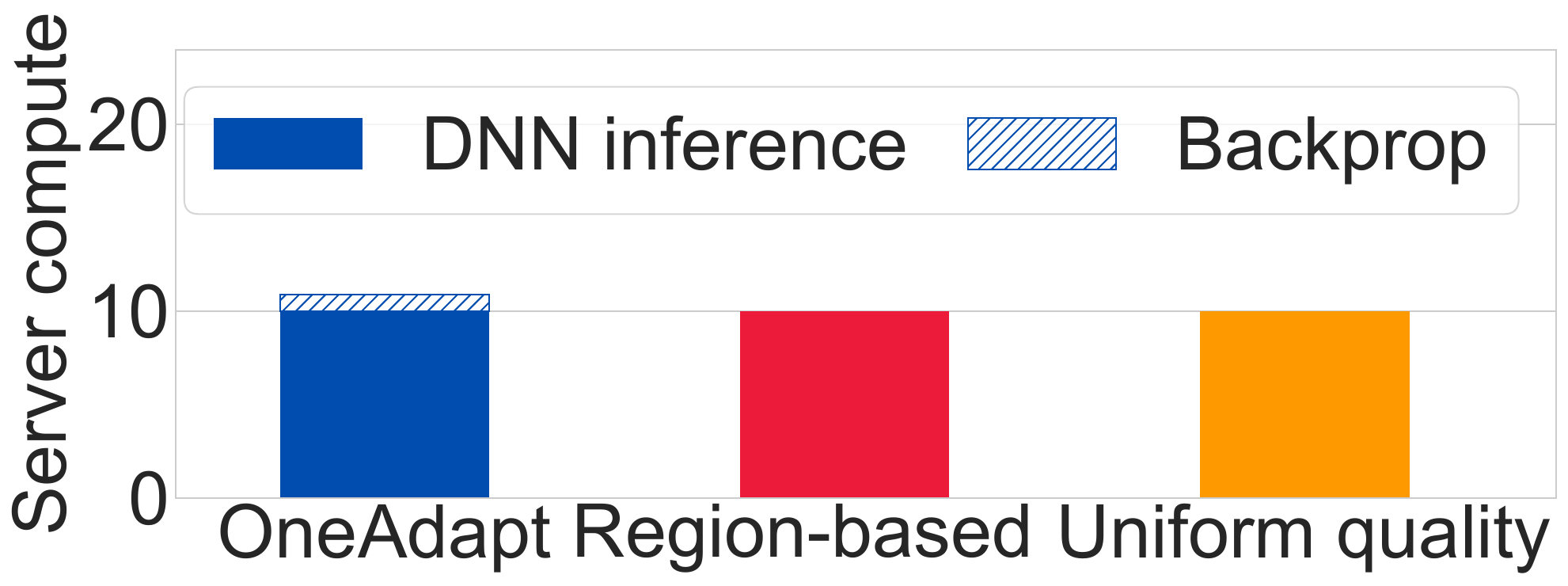} 
        \label{subfig:serv5}
    }
    
    \subfloat[][Pipeline \circled{e}\circled{f}\circled{g}  GPU comp.]
    {
        \includegraphics[width=0.49\columnwidth]{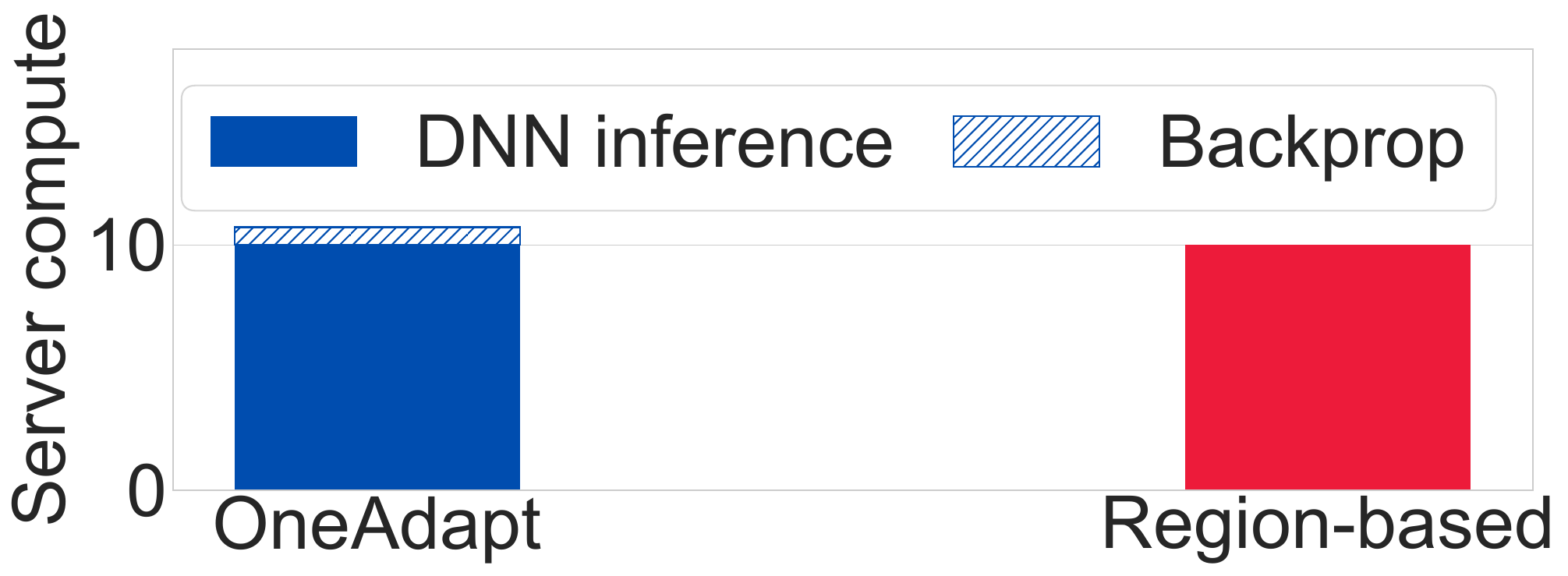} 
        \label{subfig:serv4}
    }

    \tightcaption{
    Comparing the server-side compute overheads of \name against the baselines (measured by the number of frames inferred by the server per second).}
    \label{fig:overhead}
\end{figure}

\mypara{Better trade-off between accuracy and resource}
We show that \name achieves a better trade-off between accuracy and resource usage across 9 different pipelines in Figure~\ref{fig:eval-overall-resource-consumption}.
We see that across these applications, \name achieves 15-59\% resource usage reduction compared to the baselines without decreasing inference accuracy, or improves the accuracy by 1-5\% without inflating resource usage.
We highlight that in Figure~\ref{subfig:compute-improvement}, a profiling-based baseline (Chameleon) with 8x more GPU compute than \name still has a sub-optimal trade-off between resource usage and accuracy.

\mypara{Adaptation behavior of \name}
We compared the adaptability of \name against two baselines (heuristics-based and profiling-based) using the audio-to-text pipeline (pipeline~\circled{h}).
As shown in Figure~\ref{fig:adaptivity}, up to the 42$^{nd}$ second, there is no human voice and all methods use a low audio sampling rate while maintaining 100\% accuracy. 
At the 42$^{nd}$ second, the accuracy of all methods drops due to the continued use of the low sampling rate.
\name then identifies a large \accgrad and promptly raises the sampling rate, achieving 100\% accuracy at the 45$^{th}$ second.
In contrast, the profiling baseline persists with the outdated low  sampling rate until its profiling completes at the 47$^{th}$ second. 
The voice-detection heuristics, being overly cautious, uses a conservative sampling rate, resulting in suboptimal accuracy.

\mypara{More knobs, more gain}
We show that \name achieves higher bandwidth reduction\footnote{We define bandwidth reduction as the bandwidth usage of the region-based approach, divided by the bandwidth usage of \name when \name has higher accuracy} compared to the baseline in pipeline~\circled{e}.
In Figure~\ref{fig:fine-grained-knobs}, encoding qualities are assigned to 4, 16, 64 spatial blocks, resulting in 4, 16, 64 knobs to adapt.
We show that the bandwidth reduction of \name grows larger when there are more configuration knobs.
This is because \name near-optimally handles more knobs without adding GPU computation (since \name only runs one backpropagation, regardless of the number of knobs) or CPU computation (by using the optimization for non-overlapping knobs in \S \ref{subsec:cpu-overhead-reduction}).
However, the heuristics encode similar areas in high quality regardless of number of knobs and thus cannot significantly improve the resource--accuracy trade-off when there are more knobs.

\mypara{Overhead of \name}
We measure the sensor-side CPU computation and the server-side GPU computation of \name\footnote{Note that the bandwidth overhead of streaming \dnngrad from the server to the client is negligible as it contains >7000x less amount of data after sampling and compression of \dnngrad.} using the CPU computation divided by the CPU computation of processing one frame (or 1/10 worth of data in other data formats), and the GPU computation divided by the GPU computation of analyzing one frame.
We mark the adaptation overhead of \name in the hatched area.
From Figure~\ref{fig:overhead}, we see that the server-side GPU computation of \name is comparable to or lower than the baselines, and the adaptation overhead of \name is negligible.

That said, the sensor-side CPU overhead of \name is high.
Though \name has equal or lower CPU overhead in pipeline \circled{c}\circled{h} than the baselines,
\name needs to encode the input data 3 times (in pipeline \circled{a}\circled{d}\circled{e}\circled{f}\circled{g}\circled{i}, even 4 times in pipeline\circled{b}), resulting in more CPU computation overhead than the baselines.
\review{
One may worry that \name imposes too much CPU overhead on the sensor that may exceed the sensor-side computation capability.
To answer this question, we measure the encoding speed on an Intel Xeon 4100 Silver CPU and find that it has enough compute to encode 103.2 video frames per second, which translates to encoding the input data 10 times (as the input data is 10FPS) and is sufficient for \name in our evaluation setup.}{This addresses Reviewer B's concern on CPU overhead.}



\review{\mypara{Effectiveness of GPU overhead reduction}
\name introduces two optimizations (\S\ref{subsec:saliency_sampling}) to reduce the GPU overhead caused by backpropagation: removing unneeded computation and \dnngrad reusing.
Figure~\ref{fig:GPU-overhead-reduction} tests how these two techniques reduce the backpropagation overhead of \name in terms of GPU runtime and GPU memory on pipeline~\circled{a}, on a 10-frame video chunk.
\name reduces the GPU runtime overhead of backpropagation by 87\% and the GPU memory overhead by 12\%.
Though removing unneeded computation does not change the \dnngrad, \dnngrad reusing does and may reduce the accuracy or increase the resource usage of \name.
To address this concern, we show that in Figure~\ref{fig:reduce-overhead-min-impact}, \dnngrad reusing has negligible impact on the accuracy and resource usage of \name.}
{This addresses Reviewer C's comment on the ablation study.}




\tightsection{Related work}
\label{sec:related}

\mypara{ML application systems}
Almost each ML application, from object detection~\cite{dds,eaar,vigil,glimpse,reducto,noscope,blazeit,mcdnn,accmpeg,chameleon,awstream,casva,filterforward,fischer2021saliency} to segmentation~\cite{dds,eaar,reducto,accmpeg}, has seen recent efforts towards efficient systems that with high inference accuracy and reduced resource usage in compute (model inference) and bandwidth/storage (moving data from sensors/sources to the DNN model).
They commonly entail various configurations that heavily influence resource usage and/or inference accuracy. 
For instance, Chameleon~\cite{chameleon} incorporates only two video coding parameters (resolution and frame rate), and one parameter that chooses between DNNs.
For instance, Elf~\cite{elf} proposes a new type of knob that partitions the video into different slides and distributes them to a different server for faster inference. 
Recent work~\cite{ekya,noscope,mcdnn} expands system designs with online model training as another configuration, in order to handle the drift of the input content. 

Instead of proposing a new system, we design \name to better adapt a range of existing configurations. 
That said, \name as-is does not support for {\em all} configurations, such as those that modify the analytical DNN model itself. 

\mypara{Adaptation in video analytics systems}
To cope with dynamic video input and resource availability, video analytics systems need to timely and optimally set various configurations~\cite{awstream,chameleon,accmpeg,dds,casva,vstore,videostorm,ekya,wang2022minimizing,xu2022litereconfig}.
Two general approaches exist.
One relies on (offline or periodic) profiling to search the configuration space for the optimal configuration~\cite{awstream} and leveraging the spatial-temporal locality of the input content to reduce the profiling frequency~\cite{chameleon}.
However, as elaborated in~\S\ref{subsec:configuration-adaptation}, the high overhead of such profiling makes frequent adaptation infeasible.
On the other hand, many heuristic-based solutions forgo profiling and instead select configurations based on historical DNN outputs~\cite{dds,eaar,elf} (including the intermediate outputs) or by analyzing the input data using cheap models~\cite{accmpeg,casva,vigil,filterforward,reducto,wang2022minimizing}.
This approach adapts quickly, but as shown in \S\ref{subsec:configuration-adaptation}, this approach generally sacrifices optimality and has low accuracy.

Different from prior work, \name harnesses the differentiability of DNN for configuration adaptation.
We use the differentiability of the DNN to cheaply calculate \outputgrad on all knobs with no extra inference and with constant GPU overhead, allowing \name to frequently adapt to a near-optimal configuration.

\tightsection{Limitation}
\label{sec:limitation_section}

Though \name can optimize a wide range of knobs in streaming media analytics, we have not tested if \name will work for applications beyond streaming media analytics (such as generative tasks), or knobs that alter the final DNN itself (\eg DNN selection~\cite{chameleon} and DNN customization~\cite{ekya}).

We show that the gradient-ascent strategy of \name can adapt the configurations when the content of input data changes frequently (\eg in autonomous driving scenarios).
However, this strategy may be sub-optimal when the input content changes too fast (\eg in car racing scenarios).


\review{
When handling more knobs, though the GPU overhead of \name does not inflate (since \name only runs one backpropagation), the CPU overhead of \name still linearly increases for those knobs that the CPU overhead optimization (\S\ref{subsec:cpu-overhead-reduction}) do not apply, constraining the maximum number of knobs that \name can tune.
}{This addresses Reviewer B's concern on CPU overhead.}

Also, the GPU memory overhead of \name (introduced by running backpropagation) is not negligible. That said, this overhead can be efficiently optimized by gradient checkpointing~\cite{gradient-checkpointing-1,gradient-checkpointing-2}, gradient compression~\cite{gradient-compression-1,gradient-compression-2} and reversible neural networks~\cite{reversible-1,reversible-2} and we leave optimizing this overhead to future work.







\tightsection{Conclusion}
\name addresses a common need of DNN-based applications to timely adapt key configurations over time. 
The key insight is to harness the differentiability of most DNN models, which allows 
precise estimation of the gradient of accuracy with respect to all configuration knobs with one backpropagation operation.
\name's improvement (in lower resource usage, or higher accuracy, or both) is validated in four applications, five types of configuration knobs, and five types of streaming media.


%




\bibliographystyle{plain}
\bibliography{reference}

\appendix

\section{Notation summarization}

Table~\ref{tab:notations} summarizes the notation used in our paper.

\begin{table}[]
    \scriptsize
    \centering
    \setlength\tabcolsep{0.3em}
    \begin{tabular}{|m{0.13\columnwidth}|m{0.4\columnwidth}|m{0.35\columnwidth}|}
    \hline
    {\bf Notation}                                                                                              & {\bf Definition}                                                                                                                                                                                                                                 & {\bf Example}                                                                                                         \\ \hline \hline
    $n$                                                                                                   & Number of knobs                                                                                                                                                                                                                            & $n=2$                                                                                                           \\ \hline
    $t$                                                                                                   & $t^{\textrm{th}}$ adaptation interval (by default, each interval is 1 second)                                                                                                                                                                                         & $t=1$                                                                                                           \\ \hline
    $T$                                                                                                   & Total number of adaptation intervals                                                                                                                                                                                                       & $T=60$                                                                                                          \\ \hline
    $\Config_t$                                                                                           & Configuration: a vector of knobs with their selected values at interval $t$                                                                                                                                                                & $\Config_1=$(frame rate=10, resolution=480p)                                                                    \\ \hline
    $\Knob_{i,t}$                                                                                         & The value of $i^{\textrm{th}}$ knob in configuration $\Config_t$                                                                                                                                                                                    & $\Knob_{2,1}$=480p                                                                                              \\ \hline
    $\Delta \Knob_i$                                                                                      & A small increase on the $i^{\textrm{th}}$ knob                                                                                                                                                                                                      & $\Delta \Knob_2$=120p                                                                                           \\ \hline
    $\Data_t$                                                                                             & input data at interval t                                                                                                                                                                                                                   &                                                                                                                 \\ \hline
    \begin{tabular}[c]{@{}l@{}}$\Resource(\Config_t; \Data_t)$\\ or\\ $\Resource(\Config_t)$\end{tabular} & Resource usage of config $\Config_t$ under $\Data_t$. We omit $\Data_t$ for simplicity                                                                                                                                                     &     $\Resource(\Config_1)=$5Mbps                                                                                                            \\ \hline
    \begin{tabular}[c]{@{}l@{}}$Res(\Config_t; \Data_t)$\\ or\\ $Res(\Config_t)$\end{tabular}             & Inference results of config $\Config_t$ under $\Data_t$. We omit $\Data_t$ for simplicity.                                                                                                                                                 & 
    \begin{tabular}[c]{@{}l@{}}$Res(\Config_1)=\{$\\
    (obj1, ``car'', score: 0.2),\\
    (obj2, ``bike'', score: 0.8)\\ $\}$
    \end{tabular} 
     \\ \hline
    $e$                                                                                                   & An element in the inference results (\eg a detected object). Each element is associated with a confidence score.                                                                                                                           & \begin{tabular}[c]{@{}l@{}}$e=$\\
    (obj1, ``car'', score: 0.2)
    \end{tabular}                                                                 \\ \hline
    $\theta$                                                                                              & Confidence threshold                                                                                                                                                                                                                       & $\theta=0.5$ (default)                                                                                                   \\ \hline
    \begin{tabular}[c]{@{}l@{}}$\Input(\Config_t; \Data_t)$\\ or\\ $\Input(\Config_t)$\end{tabular}         & DNN input generated by configuration $\Config_t$ using input data $\Data_t$. We omit $\Data_t$ for simplicity.                                                                                                                                           & ~                                                                                         \\ \hline
    \begin{tabular}[c]{@{}l@{}}$\Output(\Config_t; \Data_t)$\\ or\\ $\Output(\Config_t)$\end{tabular}         & Output utility: number of above-confidence-threshold elements. We omit $\Data_t$ for simplicity.                                                                                                                                           & $\Output(\Config_1)=1$                                                                                          \\ \hline
    \begin{tabular}[c]{@{}l@{}}$Acc(\Config_t; x_t)$\\ or\\ $Acc(\Config_t)$\end{tabular}                 & Accuracy of the configuration $\Config_t$ under input data $\Data_t$, defined as the similarity between the current inference result $Res(\Config_t)$ and the inference results generated using the most resource-demanding configuration. & $Acc(\Config_1)=100\%$                                                                                          \\ \hline
    $\alpha$                                                                                              & Learning rate of \name                                                                                                                                                                                                                     & $\alpha=0.5$ (default)                                                                                                    \\ \hline
    $\lambda$                                                                                             & The hyperparameter that trade-off between accuracy and resource usage.                                                                                                                                                                     & $\lambda=1$                                                                                                  \\ \hline
    \end{tabular}
    \vspace{0.1cm}
    \tightcaption{Summary of the notations used in \name}
    \label{tab:notations}
    \end{table}

\section{Implementation of convolution}
\label{sec:backprop-cost-reduction}

Beyond the existing GPU optimization, we also provide another optimization that further reduces the GPU computation and memory cost of backpropagation.
We contrast the implementation of the convolution operator between normal backpropagation and the backpropagation in \name in Algorithm~\ref{alg:conv} and Algorithm~\ref{alg:oneadapt-conv}.

\begin{algorithm}
\caption{Convolution}
\begin{algorithmic}
\Function{forward}{cache, input, kernel}
    \State output $\gets$ {\sf convolution}(input, kernel)
    \State cache.push(input, kernel)
\EndFunction
\Function{backward}{cache, outputGrad}
    \State input, kernel $\gets$ cache.pop()
    \State inputGrad $\gets$ {\sf convGradInput}(gradOutput, kernel)
    \State kernelGrad $\gets$ {\sf convGradKernel}(outputGrad, input)
\EndFunction
\end{algorithmic}
\label{alg:conv}
\end{algorithm}

\begin{algorithm}
\caption{Optimized Convolution}
\begin{algorithmic}
\Function{forward}{cache, input, kernel}
    \State output $\gets$ {\sf convolution}(input, kernel)
    \State cache.push({\color{red} kernel.abs().mean(dim=`channel')})
\EndFunction
\Function{backward}{cache, outputGrad}
    \State kernel $\gets$ cache.pop()
    \State inputGrad $\gets$ {\sf convGradInput}(outputGrad, kernel)
    \State \st{kernelGrad $\gets$ \sf{convGradKernel}(outputGrad, input)}
\EndFunction
\end{algorithmic}
\label{alg:oneadapt-conv}
\end{algorithm}

The intuition is that \name focuses on the \dnngrad of different spatial areas instead of different channels between different channels in the DNN input (\eg the RGB color channel).
We empirically find that this optimization has little impact on the bandwidth--accuracy trade-off of \name.

\section{Reusing \dnngrad}

\begin{figure}[t]
    \centering
   
        \includegraphics[width=0.45\columnwidth]{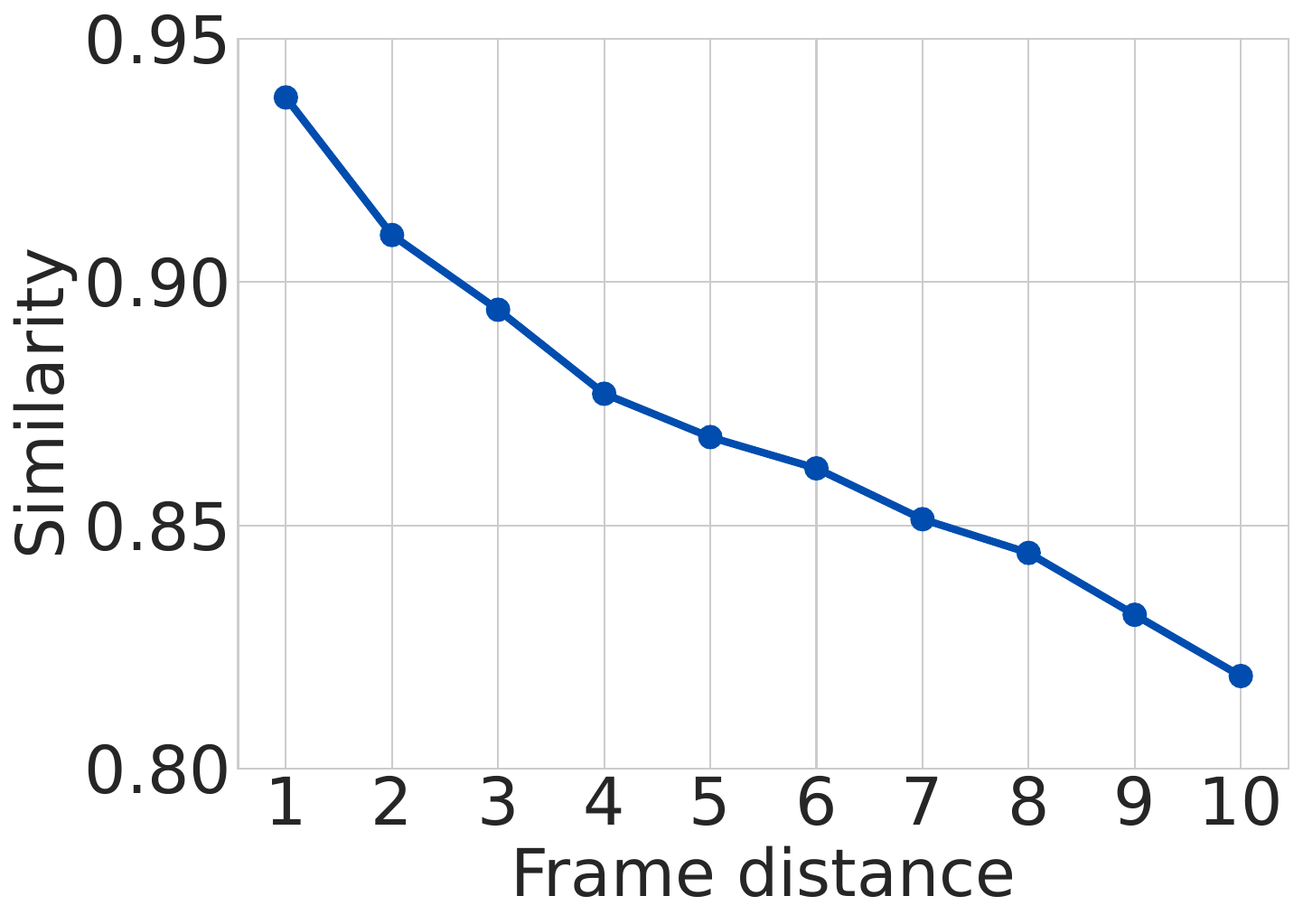} 
        \label{subfig:1}


    \tightcaption{The similarity of the saliency remains higher than 80\% when the frame distance is less than 10.}
    \label{fig:reusing-saliency}
\end{figure}

The intuition of reusing \dnngrad is that: although the exact value of \dnngrad may vary across different frames, the spatial areas that have high \dnngrad tend to be stable within several consecutive frames.
To verify this intuition, we observe how the similarity (we use cosine similarity) between the saliency of two frames changes with respect to the distance between these two frames
(measured by the absolute difference of the video frame id) on a three-second video in our dataset. 
As shown in Figure~\ref{fig:reusing-saliency}, we observe that this similarity decreases when the frame distance becomes larger, but the similarity is still greater than 80\% when the frame distance is less than 10.

\section{Summarizing the dataset}

Table~\ref{tab:dataset} summarizes the dataset used in \name.

\begin{table}[]
\footnotesize
\begin{tabular}{|l|l|l|l|l|}
\hline
Dataset                                                                                         & \begin{tabular}[c]{@{}l@{}}Streaming\\media type\end{tabular}            & \begin{tabular}[c]{@{}l@{}}\scriptsize \#videos/\\ \scriptsize\#audios\end{tabular} & \begin{tabular}[c]{@{}l@{}}Total\\ length\end{tabular} & Description                                                                               \\ \hline
\sf{Traffic}~\cite{googleOneAdaptDriving}               & \multirow{3}{*}{RGB} & 5                                                                       & 5min                             & {\scriptsize Traffic camera footage}                                                     \\ \cline{1-1} \cline{3-5} 
\sf{Downtown}~\cite{googleOneAdaptDriving}             &                      & 10                                                                      & 20min                                                  & {\scriptsize Driving in downtwon}                                                                       \\ \cline{1-1} \cline{3-5} 
\sf{Country}~\cite{googleOneAdaptDriving}               &                      & 8                                                                       & 18min                                                  & {\scriptsize Driving in countryside}                                                                    \\ \hline
\multirow{3}{*} {\scriptsize PKU-MMD~\cite{pku-mmd}}                                & RGB                  & 10                                                                      & 20min                                                  & \multirow{3}{*}{\begin{tabular}[c]{@{}l@{}}Human moving\\ in a static scene\end{tabular}} \\ \cline{2-4}
                                                                                                & Depth                & 10                                                                      & 20min                                                  &                                                                                           \\ \cline{2-4}
                                                                                                & Infrared             & 10                                                                      & 20min                                                  &                                                                                           \\ \hline
KITTI~\cite{kitti}                                                       & LiDAR                & 6                                                                       & 4min                                                   & {\scriptsize Driving on city streets}                                                                   \\ \hline
\begin{tabular}[c]{@{}l@{}}Google\\AudioSet~\cite{audioset}\end{tabular} & Audio                & 200                                                                     & 33min                                                  & \begin{tabular}[c]{@{}l@{}}\scriptsize Advertisement and \\\scriptsize leisure activities\end{tabular}                                                     \\ \hline
\end{tabular}
\vspace{0.2cm}
\tightcaption{Summary of our dataset.}
\vspace{-0.2cm}
    \label{tab:dataset}
\end{table}

\begin{figure}[t]
    \centering
    \includegraphics[width=0.45\columnwidth]{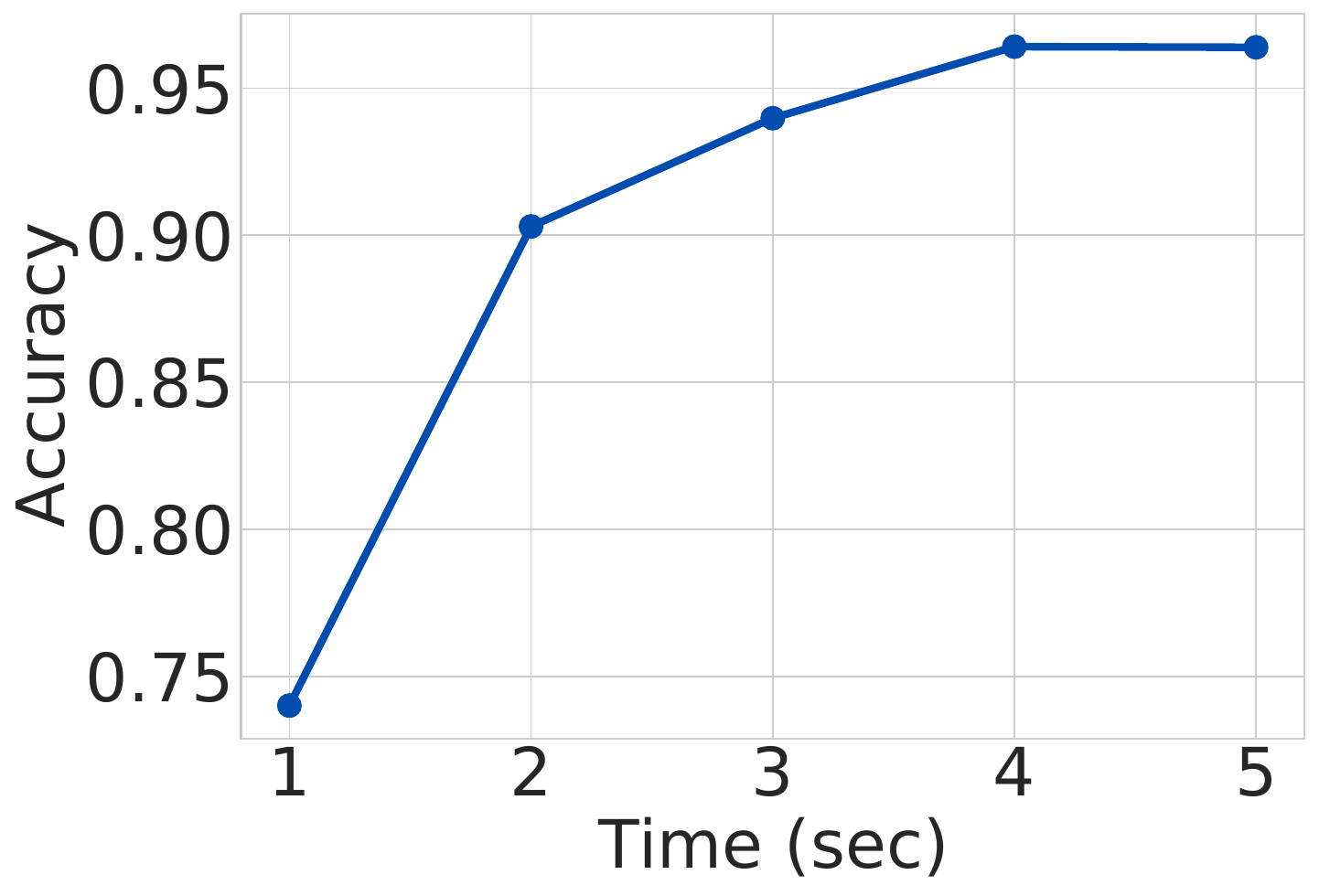}
    \tightcaption{
    Empirically \name converges within 3-5 seconds.
    }
    \label{fig:converge-fast}
\end{figure}

\section{Theoretical correlation between \outputgrad and \accgrad}
\label{sec:proof}

In this section, we prove the correlation between \accgrad and \outputgrad by treating them as the analytical gradient (so the chain rule of derivative holds).
We are actively working on refining the proof to extend it to numerical gradient case~\cite{OneAdapt-proof}.

\mypara{Definitions}
For a given input, we let $Res(\Config)$ denote the output under a configuration $\Config$.
We define its accuracy as 
\begin{equation}
Acc(\Config) = \sum_{e\in Res(\Config)} f(e)C(e),
\label{eq:accuracy-definition}
\end{equation}
and the output utility as 
\begin{equation}
\Output(\Config) = \sum_{e \in Res(\Config)} f(e).
\label{eq:outputgrad}
\end{equation}
where $f(e)$ is a differentiable function that returns a value close to $1$ when $e$ is confidently detected (\ie the confidence score of $e$ is greater than the confidence threshold $\theta$) and close to $-1$, otherwise\footnote{Note that in our evaluation we use $f(e) = 2Sigmoid\left(20 (e.score-\theta)\right)-1$ (where $e.score$ is the confidence score of $e$ and $\theta$ is the confidence threshold), but the choice of $f$ does not affect the validity of our proof.
}, and $C(e)=1$ if $e$ is confidently detected under the most expensive configuration, and $C(e)=-1$ otherwise.

\mypara{Theorem}
For any knob $\Knob$ in configuration $\Config$, we have
$$
\underbrace{\frac{\partial Acc(\Config)}{\partial \Knob}}_{\text{\sf \footnotesize \textcolor{black!85}{\accgrad}}} = \underbrace{
\left|
\frac{\partial \Output(\Config)}{\partial \Knob}
\right|
}_{{\text{\sf \footnotesize \textcolor{black!85}{\outputgrad}}}}.
$$ 
under the following assumptions:
\begin{packedenumerate}
\item 
For each $\Knob$ in the configuration $\Config$ and any element $e$ in $Res(\Config)$, we have:
\begin{@empty}
\begin{equation}
\frac{\partial f(e)C(e)}{\partial \Knob} \geq 0.
\label{eq:acc-grad-property}
\end{equation}
\end{@empty}

\item 
For any element $e$ in $Res(\Config)$, there exists a binary matrix $M(e)$ s.t.
\begin{@empty}
\begin{equation}
M(e) \times \frac{\partial \Output(\Config)}{\partial \Input}=
M(e) \times \frac{ \partial  \sum_{e'\in Res(\Config)} f(e')}{\partial \Input} = \frac{\partial f(e)}{\partial \Input},
\label{eq:mask-property}
\end{equation}
\end{@empty}
where $\times$ means element-wise multiplication and $\Input$ refers to DNN input. 

\item 
Finally, we assume
\begin{@empty}
\begin{equation}
\left|
\frac{\partial \Output(\Config)}{\partial \Input}
\otimes 
\frac{\partial \Input}{\partial \Knob}
\right|
=
\left|
\frac{\partial \Output(\Config)}{\partial \Input}
\right|
\otimes
\left|
\frac{\partial \Input}{\partial \Knob}
\right|
\label{eq:non-negativity}
\end{equation}
\end{@empty}

\end{packedenumerate}

\mypara{Proof}

\begin{@empty}
\footnotesize
\begin{alignat*}{2}
&~~~~\frac{\partial Acc(\Config)}{\partial \Knob} \\
&= \frac{\partial \sum_{e\in Res(k)}f(e)C(e)}{\partial \Knob}
&\quad &\text{(definition of accuracy)}\\
&=\sum_{e\in Res(k)}\frac{\partial f(e)C(e)}{\partial \Knob}
&\quad &\text{(linearity of derivative)}\\    
&=\sum_{e\in Res(k)}\left|\frac{\partial f(e)C(e)}{\partial \Knob}\right|\
&\quad &\text{(equation~\ref{eq:acc-grad-property})}\\
&=\sum_{e\in Res(k)}\left|\frac{\partial f(e)}{\partial \Knob}\right|\
&\quad &(|C(e)|=1)\\
&=\sum_{e\in Res(k)}\left|\frac{\partial f(e)}{\partial \Input}\otimes \frac{\partial \Input}{\partial \Knob}\right|
&\quad &\text{(chain rule,$\Input$ is DNN input)}\\   
&=\sum_{e\in Res(k)}\left|M(e)\times \frac{\partial \Output(\Config)}{\partial \Input}\otimes \frac{\partial \Input}{\partial \Knob}\right|
&\quad &\text{(equation~\ref{eq:mask-property})}\\
&=\sum_{e\in Res(k)}M(e)\times \left|\frac{\partial \Output(\Config)}{\partial \Input}\otimes \frac{\partial \Input}{\partial \Knob}\right|
&\quad &\text{($M(e) $is binary matrix)}\\
&=\sum_{e\in Res(k)}M(e)\times \left|\frac{\partial \Output(\Config)}{\partial \Input}\right|\otimes \left|\frac{\partial \Input}{\partial \Knob}\right|
&\quad &\text{(equation~\ref{eq:non-negativity})}\\
&=\left(\sum_{e\in Res(k)}M(e)\times \left|\frac{\partial \Output(\Config)}{\partial \Input}\right|
\right)
\otimes 
\left|\frac{\partial \Input}{\partial \Knob}\right|
&\quad &\text{(linearity of inner product)}\\
&=\left(
\left(\sum_{e\in Res(k)}M(e)\right)\times \left|\frac{\partial \Output(\Config)}{\partial \Input}\right|
\right)
\otimes 
\left|\frac{\partial \Input}{\partial \Knob}\right|
&\quad &\text{(distributivity)}\\
&=\left|\left(\sum_{e\in Res(k)}M(e)\right)\times \frac{\partial \Output(\Config)}{\partial \Input}\right|
\otimes 
\left|\frac{\partial \Input}{\partial \Knob}\right| 
&\quad &\text{($M(e)$ is non-negative)}\\
&=\left|\sum_{e\in Res(k)}M(e)\times \frac{\partial \Output(\Config)}{\partial \Input}\right|
\otimes 
\left|\frac{\partial \Input}{\partial \Knob}\right| 
&\quad &\text{(distributivity)}\\
&=\left|\sum_{e\in Res(k)}\frac{\partial f(e)}{\partial \Input}\right|
\otimes 
\left|\frac{\partial \Input}{\partial \Knob}\right|
&\quad &\text{(equation~\ref{eq:mask-property})}\\
&=\left|\frac{\partial \sum_{e\in Res(k)} f(e)}{\partial \Input}\right|
\otimes 
\left|\frac{\partial \Input}{\partial \Knob}\right|
&\quad &\text{(linearity of derivatives)}\\
&=\left|\frac{\partial \Output(\Config)}{\partial \Input}\right|
\otimes 
\left|\frac{\partial \Input}{\partial \Knob}\right|
&\quad &\text{(definition of \Output(\Config), we}\\
&~&~&\text{estimates \accgrad this way)}\\
&=\left|\frac{\partial \Output(\Config)}{\partial \Input}
\otimes 
\frac{\partial \Input}{\partial \Knob}\right|
&\quad &\text{(equation~\ref{eq:non-negativity})}\\
&=\left|\frac{\partial \Output(\Config)}{\partial \Knob}\right|
&\quad &\text{(chain rule)}\\
\end{alignat*}
\end{@empty}







Empirically, we show in Figure~\ref{fig:accgrad-outputgrad-cosine-similarity}, the cosine similarity between \accgrad and \outputgrad still remains over 0.91, indicating that they still have high correlation.




\end{document}